\documentclass[10pt,twocolumn,letterpaper]{article}

\usepackage[pagenumbers]{cvpr} 
\usepackage{subcaption}
\usepackage{amsmath}
\usepackage{graphicx}
\usepackage{booktabs}
\usepackage{multirow}
\usepackage[table]{xcolor}
\definecolor{teal}{RGB}{0,128,128}
\newcommand{\method}{AVION}
\usepackage{amssymb}
\usepackage{amsmath}
\usepackage{booktabs}
\usepackage{multirow}
\usepackage{graphicx}
\usepackage[table]{xcolor} 
\usepackage{colortbl}      
\usepackage{makecell}      
\usepackage[table]{xcolor}   
\usepackage{booktabs}        
\usepackage{makecell}
\usepackage{adjustbox}

\usepackage[hyphens]{url}            %
\definecolor{RowBlue}{HTML}{E8F1FE} 
\definecolor{RowRed}{HTML}{FDE9E9}

\newcolumntype{L}[1]{>{\raggedright\arraybackslash}p{#1}}
\newcolumntype{C}[1]{>{\centering\arraybackslash}p{#1}}
\newcolumntype{R}[1]{>{\raggedleft\arraybackslash}p{#1}}
\definecolor{TableGray}{gray}{0.9}
\definecolor{HighlightPurple}{RGB}{230, 230, 250}

\usepackage{microtype}
\usepackage[accsupp]{axessibility}

\renewcommand{\paragraph}[1]{\vspace{1.2mm}\noindent\textbf{#1}}

\definecolor{cvprblue}{rgb}{0.21,0.49,0.74}
\usepackage[pagebackref,breaklinks,colorlinks,allcolors=cvprblue]{hyperref}

\def\paperID{1469} 
\def\confName{CVPR}
\def\confYear{2026}

\title{\method{}: Aerial Vision–Language Instruction from \\ Offline Teacher to Prompt-Tuned Network}

\author{
Yu Hu$^{1}$\thanks{Both authors contributed equally to this research.}\quad
Jianyang Gu$^{2}$\footnotemark[1]\quad
Hao Liu$^{1}$\quad
Yue Cao$^{1}$\quad
Jozsef Hamari$^{3}$\quad
Zheng Liu$^{1}$\thanks{Corresponding authors.}\quad
Mohsen Zardadi$^{3}$\footnotemark[2]\\
$^{1}$The University of British Columbia, Okanagan, Kelowna, BC, Canada\\
$^{2}$The Ohio State University, Columbus, OH, USA\\
$^{3}$TerraSense Analytics, Kelowna, BC, Canada\\
{\tt\small yu.hu@ubc.ca, gu.1220@osu.edu, hao.liu@ubc.ca, yue.cao@ubc.ca}\\
{\tt\small \{jozsef.hamari, mohsen.zardadi\}@terrasense.ca, zheng.liu@ubc.ca}\\
}

\begin{document}
\maketitle


\begin{abstract}
Adapting vision-language models to remote sensing imagery remains challenging due to two key factors: limited semantic coverage in textual representations and insufficient adaptability of visual features. These issues are particularly significant in aerial scenes, which involve various visual appearances and fine-grained object distinctions. 
We propose \method{}, a knowledge distillation framework tailored for remote sensing adaptation of vision-language models. The teacher module constructs semantically rich textual prototypes by collecting descriptions from a large language model and verifying validity using remote sensing image features. The student module integrates lightweight and learnable prompts into both vision and language encoders, guided by the teacher to align embeddings and their cross-modal relationships. Once trained, the student operates independently during inference.
Experiments on six optical remote sensing benchmarks show that \method{} improves few-shot classification and base-class accuracy without degrading generalization to novel categories. It also enhances mean recall for cross-modal retrieval, with minimal additional trainable parameters. Code is publicly available at \href{https://github.com/yuhu990424/AVION}{https://github.com/yuhu990424/AVION}.
\end{abstract}    
\section{Introduction}
\label{sec:intro}

Domain-specific vision--language models (VLMs) adapt rich knowledge of general-purpose foundation models to specific applications~\cite{awais2025foundation,huang2025survey,yin2024survey}.
For example, RemoteCLIP~\cite{liu2023remoteclip} and GeoRSCLIP~\cite{zhang2024rs5m} have achieved state-of-the-art performance in remote sensing (RS) after extensive training. However, it remains a significant challenge to efficiently adapt these  large-scale foundation models to new scenarios.
Fully fine-tuning the entire backbone is an ideal option to boost the performance on new tasks, but it is computationally expensive and slow. This approach is often infeasible in real-world scenarios that demand rapid deployment while having only scarce labels and limited computational resources~\cite{gao2024clip,zhang2022tipadapter}.

\begin{figure}[t]
    \centering
    \includegraphics[width=0.95\linewidth]{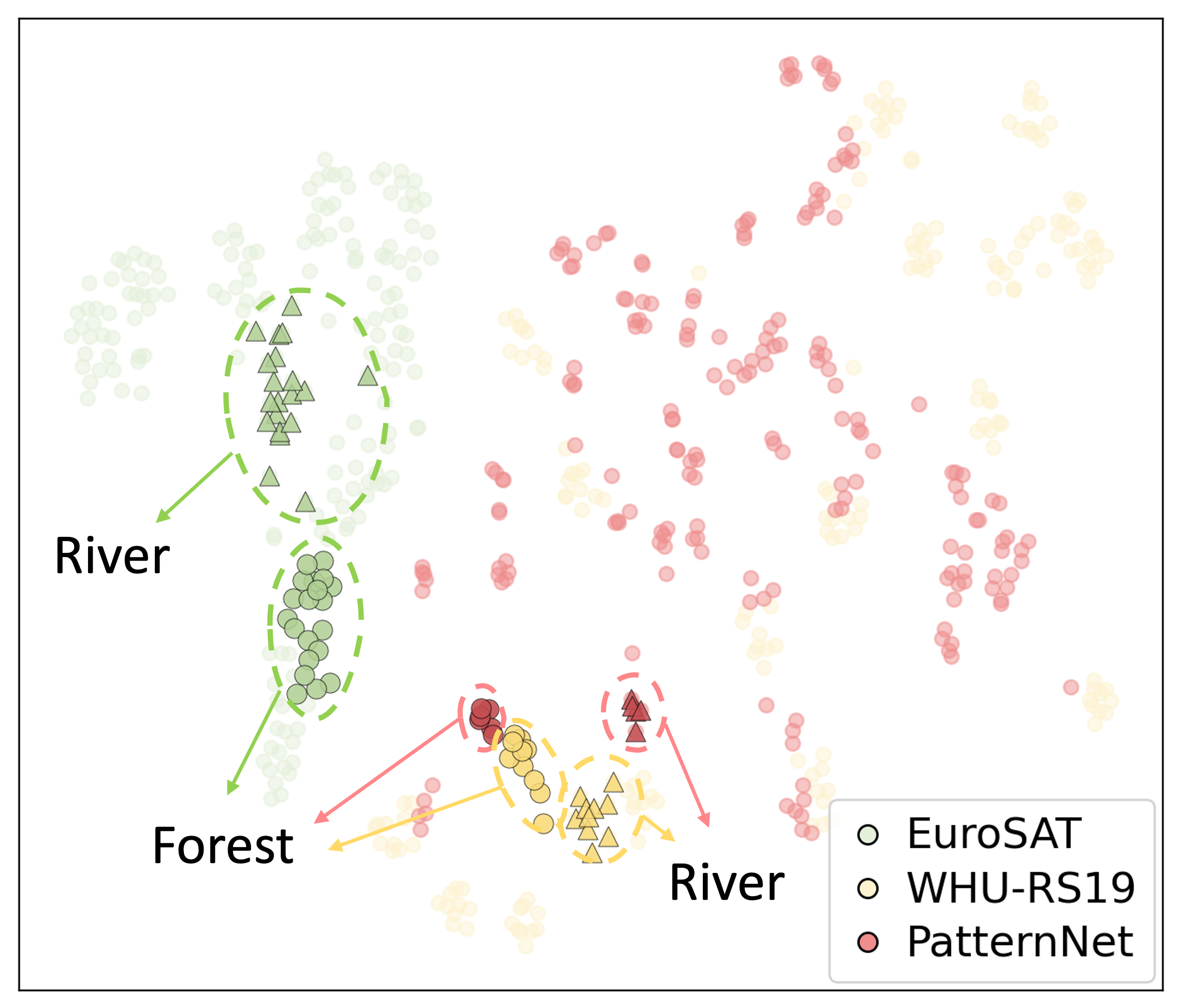}
\vspace{-4pt}
    \caption{\textbf{The t-SNE visualization of visual embeddings from different datasets.} The same class demonstrates large variations across datasets. However, only class names are provided in the datasets, which limits multimodal alignment. }
    \label{fig:intro}
\vspace{-8pt}
\end{figure}

Parameter-efficient fine-tuning (PEFT) emerges as a lightweight adaptation solution~\cite{zhou2022coop, khattak2023maple}. These methods only update a small set of parameters, leaving the backbone of foundation models frozen. While efficient, existing PEFT methods remain sub-optimal for RS. As illustrated in Fig.~\ref{fig:intro}, the same category can have disparate visual appearances across different datasets. However, the images are only paired with the class name, which cannot effectively capture such visual variances. 
PEFT methods like CoOp~\cite{zhou2022coop} learn new tasks from simple templates (\eg, \texttt{a photo of [CLASS]}). 
As a result, the text encoder cannot adequately describe the diverse appearance patterns in RS data. We denote this limitation as \textbf{\emph{semantic poverty}}. Such a disconnection between labels and semantics also makes generalization to novel classes challenging. 

Moreover, many PEFT methods primarily update the text encoder while freezing the visual encoder. We argue that this design obscures models from capturing scale variation (\eg, large spatial context and subtle textures) and heterogeneity across sources in RS imagery. 
We refer to this limitation as \textbf{\emph{visual rigidity}}. 
Although PEFT can also act on the vision side through visual prompting~\cite{jia2022vpt} and lightweight adapters~\cite{guo2025mmrl}, they still struggle to capture RS-specific visual features without proper guidance from the textual side.


To this end, we introduce Aerial Vision–Language InstructiON (\textbf{\method{}}), a knowledge distillation framework that addresses the above issues.
A frozen large teacher model is employed to distill knowledge into a lightweight student model that is later used for inference. 
To tackle the \textit{semantic poverty}, we use a large language model (LLM) to generate semantically rich descriptions for each class.
The teacher model then aggregates these descriptions into a robust textual prototype based on its similarity with the visual embeddings of the corresponding class. 
This prototype contains much richer information compared with the class name alone. 
For \textit{visual rigidity}, we inject learnable prompts to the visual side like VPT~\cite{jia2022vpt}. These prompts give the visual encoder flexibility to handle visual variations (\eg, oblique views) in the RS scenarios.

Then, we apply distillation on the visual embeddings, textual embeddings, and their similarity logits simultaneously. 
This tri-aspect distillation helps the student acquire the visual and textual understandings, as well as the multimodal alignment structure constructed by the teacher model.
The learnable prompts for both vision and text encoders allow the student model to accumulate rich knowledge on diverse RS imagery, guided by robust textual prototypes. This flexibility maintains the generalization to unseen data. As shown in Tab.~\ref{tab:base2novel_avg}, \method{} is the only method among state-of-the-art approaches that yields higher base-to-novel accuracy than the GeoRSCLIP baseline.

Our contributions are summarized as follows:
\begin{itemize}
    \item We identify and analyze two critical limitations of previous PEFT methods for remote sensing imagery: semantic poverty and visual rigidity.
    \item We propose \method{}, a knowledge distillation framework that addresses both limitations with semantically rich textual prototypes and multimodal prompt tuning for both visual and text encoders.
    \item With a tri-aspect alignment, \method{} consistently surpasses other state-of-the-art PEFT methods in few-shot classification, base-to-novel generalization, and frozen-gallery retrieval settings.
\end{itemize}

\section{Related Work}
\label{sec:rw}

\subsection{Vision-Language Models}
VLMs align images and text in a shared embedding space and deliver strong zero-shot classification and cross-modal retrieval performances. Contrastive Language-Image Pre-training (CLIP) models train an image and a text encoder in a contrastive fashion~\cite{radford2021clip,jia2021align}. Subsequent advances have refined objectives~\cite{zhai2023siglip, tschannen2025siglip2multilingualvisionlanguage}, introduced fine-grained interactions~\cite{yao2021filip}, or explored efficient transfer recipes~\cite{zhai2022lit, cherti2023openclip}. Other architectures have explored generative/fusion modeling~\cite{li2021albef, yu2022coca} or incorporating LLMs via lightweight bridges (\eg, BLIP-2~\cite{li2023blip2}).

\subsection{Parameter-Efficient Adaptation of VLMs}
The heavy foundation models motivate PEFT methods. Among these, prompt learning has emerged as a prominent approach. Text-side prompting (\eg, CoOp~\cite{zhou2022coop}, CoCoOp~\cite{zhou2022cocoop}) optimizes learnable context tokens, while vision-side prompting (VPT~\cite{jia2022vpt}) inserts prompts into ViT layers. Multi-modal prompting (MaPLe~\cite{khattak2023maple}) introduces paired prompts for both encoders. Beyond prompting, non-prompt adapters (\eg, CLIP-Adapter~\cite{gao2024clip}, Tip-Adapter~\cite{zhang2022tipadapter}) attach lightweight heads or training-free caches. MMRL injects a learnable modality-agnostic representation space into higher layers to balance specialization and generalization~\cite{guo2025mmrl}. PromptKD uses unlabeled images for logit-based distillation from a CLIP teacher to learnable prompts~\cite{li2024promptkd}.
Prompt-CAM uses PEFT to reveal interpretable traits~\cite{chowdhury2025prompt}.
While efficient, these PEFT methods are sub-optimal in aligning diverse visual appearances with their class names in the RS scenario. \method{} tackles this challenge via introducing semantically rich textual prototypes and prompt tuning for both vision and text encoders.

\begin{figure*}[t]
    \centering
    \includegraphics[width=0.86\linewidth]{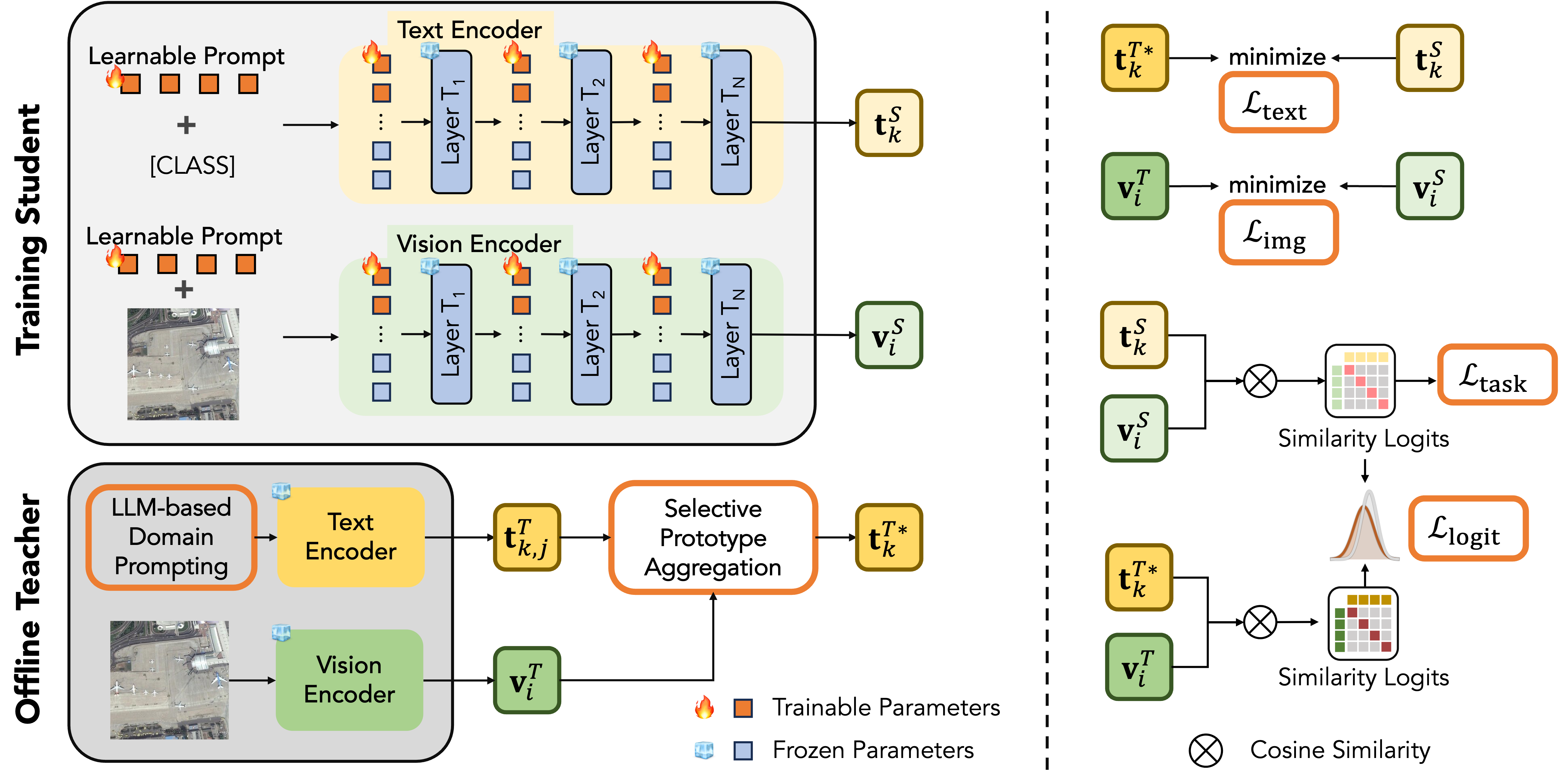}
\vspace{-6pt}
    \caption{\textbf{Overview of \method{}.}
\textbf{Upper-left (Training Student).} Learnable prompt tokens are injected into the text and vision encoders. The student outputs embeddings: ${\mathbf t}^{S}_{k}$ (student text embedding for class $k$) and ${\mathbf v}^{S}_{i}$ (visual embedding for image $i$).
\textbf{Bottom-left (Offline Teacher).} \emph{LLM-based Domain Prompting} generates multiple class-aware descriptions, which are encoded into $\mathbf t^{T}_{k,j}$. \emph{Selective Prototype Aggregation} verifies these candidates using teacher visual embeddings ${\mathbf v}^{T}_{i}$ and aggregates them into an RS-aware text prototype ${\mathbf t}^{T*}_{k}$.
\textbf{Right (Training Objectives).} Tri-aspect alignment: (i) \emph{textual alignment} $\mathcal L_{\text{text}}$ maximizes the cosine similarity between ${\mathbf t}^{S}_{k}$ and ${\mathbf t}^{T*}_{k}$; (ii) \emph{visual alignment} $\mathcal L_{\text{img}}$ aligns ${\mathbf v}^{S}_{i}$ with ${\mathbf v}^{T}_{i}$; (iii) \emph{similarity logit alignment} $\mathcal L_{\text{logit}}$ matches teacher logits $s^{T}_{i,k}=({\mathbf v}^{T}_{i})^{\top}{\mathbf t}^{T*}_{k}$ and student logits $s^{S}_{i,k}=({\mathbf v}^{S}_{i})^{\top}{\mathbf t}^{S}_{k}$ via temperature-scaled KL. A standard task loss $\mathcal L_{\text{task}}$ (cross-entropy) is applied on the student.}
    \label{fig:overview}
\vspace{-6pt}
\end{figure*}

\subsection{VLM Application in Remote Sensing}
Remote sensing (RS) images present a significant visual domain gap compared to general-purpose web data. Unlike ground-level photos, RS images are captured from top-down perspectives, with significant variations in object scale and resolution. The progress of the VLM application has followed two fronts. The first front builds RS-specific foundation models. Works like RemoteCLIP~\cite{liu2023remoteclip}, RS5M/GeoRSCLIP~\cite{zhang2024rs5m}, and LRSCLIP~\cite{chen2025lrsclip} build large-scale RS datasets to train domain-specific models. Others align images directly with geographic coordinates (\eg, GeoCLIP~\cite{vivanco2023geoclip}, SatCLIP~\cite{klemmer2025satclip}). The second front focuses on parameter-efficient RS adaptation. For classification, APPLeNet~\cite{jha2023applenet} generates image-conditioned visual prompt tokens by fusing multi-scale content features with domain style information. MVP~\cite{zhu2024mvp} integrates visual prompting into a meta-learning framework. RotCLIP~\cite{song2025rotclip} augments CLIP with a lightweight Rot-Adapter and dual textual prompts. RSPrompter~\cite{chen2024rsprompter} and RSRefSeg~\cite{chen2025rsrefseg} learn prompts for segmentation. PeftCD~\cite{dong2025peftcd} plugs LoRA and Adapter modules into Siamese backbones for change detection.
Compared with other explorations, \method{} focuses on the intrinsic visual variation of RS imagery. 
The distillation improves adaptation results under realistic settings while retaining efficiency. 

\section{Methodology}
\label{sec:method}


Compared with general web images, remote sensing (RS) imagery poses unique challenges~\cite{huo2025remote,datla2024learning}.
First, RS images present a large gap from general images, which stems from the top-down perspective. 
Second, within RS domains, the same semantic category can have drastically different visual appearances due to the variations of regions, seasons, sensor types, and resolutions. 
These challenges make existing PEFT methods inadequate for adapting foundation models to new scenarios. In this work, we propose an \method{} distillation framework for more robust adaptation performance without sacrificing efficiency. 
An overview of the proposed \method{} framework is illustrated in Fig.~\ref{fig:overview}. 



\subsection{Preliminary}
\label{subsec:notation}
Let $\mathcal{X}{=}\{(\mathbf{x}_i,y_i)\}_{i=1}^{N}$ be the training set, where the label $y_i\in\mathcal{Y}{=}\{1,\dots,C\}$. 
We employ a large teacher model in the training process to provide distillation supervision. 
Learnable deep prompts are injected into a student model whose backbone parameters are frozen~\cite{jia2022visualprompttuning}.
We denote the visual and textual embeddings encoded by the teacher model as $\mathbf{v}^T$ and $\mathbf{t}^T$, respectively. Similarly, the student's visual and textual embeddings are denoted as $\mathbf{v}^S$ and $\mathbf{t}^S$.
Logit $s_{i,k}$ represents the similarity between visual embedding $\mathbf{v}_i$ and textual embedding $\mathbf{t}_k$.
We present a summary of all the symbols and hyperparameter settings in Appendix~\ref{app:symbols}. 

As discussed, adapting VLMs to RS is challenged by two critical limitations: (1) \textit{semantic poverty}, as simple class names fail to capture fine-grained visual distinctions; (2) \textit{visual rigidity}, as the frozen visual encoder struggles with RS-specific features. There is a necessity for a framework that enriches textual supervision while enhancing visual adaptability to mitigate the gap of RS-tailored adaptation.

Therefore, we propose to design a distillation framework that efficiently adapts RS foundation models to new application scenarios. Specifically, we propose to enhance the textual prototype to incorporate richer semantic information as distillation supervision. Moreover, the proposed \method{} framework aligns both embeddings and similarity logits during distillation. 

\subsection{Textual Prototype Enhancement}
\label{subsec:teacher}
The teacher model encodes images and text labels as distillation supervisions for the student model. 
Compared with the visual embeddings conditioned on different images, the adopted RS datasets often only provide the class name without detailed descriptions. Therefore, the textual embeddings can be regarded as class-specific prototypes. The corresponding visual embeddings are optimized toward the textual prototype during distillation. It also indicates that for a dataset with $C$ classes, it will also have $C$ different textual prototypes. 
As aforementioned, the class names alone cannot provide sufficient information for the intrinsic visual variations across different RS scenarios. 
Thus, we propose incorporating rich knowledge of large language models (LLMs) to generate descriptions for RS classes. 

\begin{figure}
    \centering
    \includegraphics[width=0.8\linewidth]{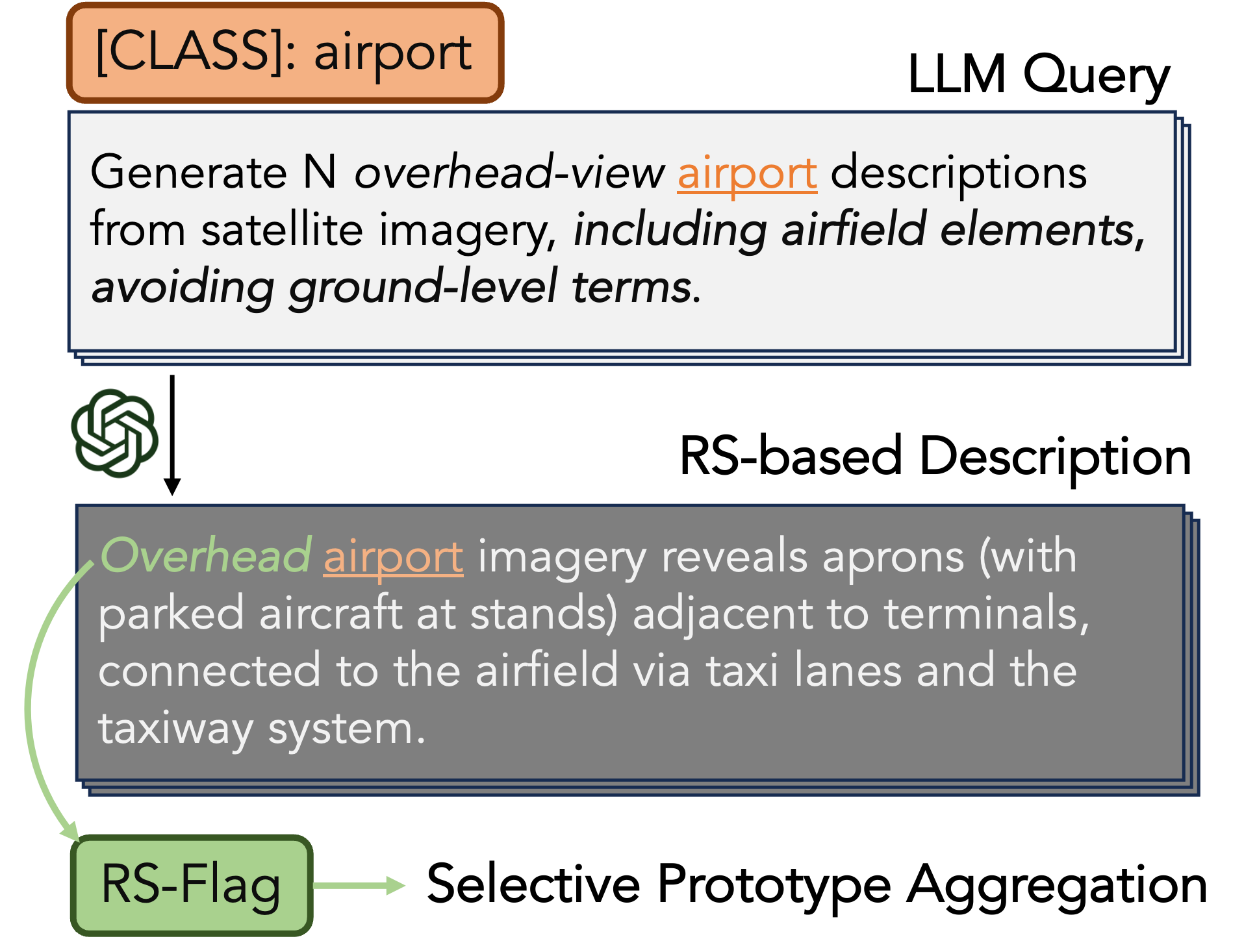}
    \vspace{-4pt}
    \caption{\textbf{LLM-based Domain Prompting.}
Given a class $k$, an RS-aware query asks the LLM to produce aerial-view descriptions. \texttt{RS-Flag} is used to examine whether the description contains RS-related tokens.}

    \label{fig:RSDP}
    \vspace{-8pt}
\end{figure}

\paragraph{LLM-based domain prompting.}
\label{subsec:rsdp}
Given a class $k\in\mathcal{Y}$, we prompt LLM to generate the corresponding descriptions in the context of RS. 
As shown in Fig.~\ref{fig:RSDP}, the generated descriptions largely improve the semantic richness of the textual supervision. 
The detailed prompt and concrete examples of these generated candidates are presented in Appendix~\ref{sec:appendix_rsdp_examples}. 

While presenting rich semantic information, LLMs may hallucinate non-visual or non-RS-related descriptions. The textual information might not be grounded in the images as well. 
Therefore, it is critical to verify the validity of these descriptions based on images in each dataset. 
We attach an \texttt{RS-Flag} to each description based on the following rules: the description should contain RS-related words and exclude RS-negative words; the description should be within a proper length; the description should include class-specific cues. Details of \texttt{RS-Flag} are in Appendix~\ref{app:srp_details}.


\paragraph{Selective prototype aggregation.}
\label{subsec:srp}
To verify the LLM candidates, we first cache the teacher visual embeddings $\mathbf v^{T}_{k,i}$ of class $k$ and obtain the visual prototype ${\widehat{\mathbf v}}^T_k$ by:
\begin{equation}
    {\widehat{\mathbf v}}^T_k=\frac{1}{|\mathcal{B}_k|}\sum_{i\in\mathcal{B}_k}\mathbf{v}^T_{k,i},
\end{equation}
where $\mathcal{B}_k$ is the sample set of the class $k$.

\begin{figure}
    \centering
    \includegraphics[width=\linewidth]{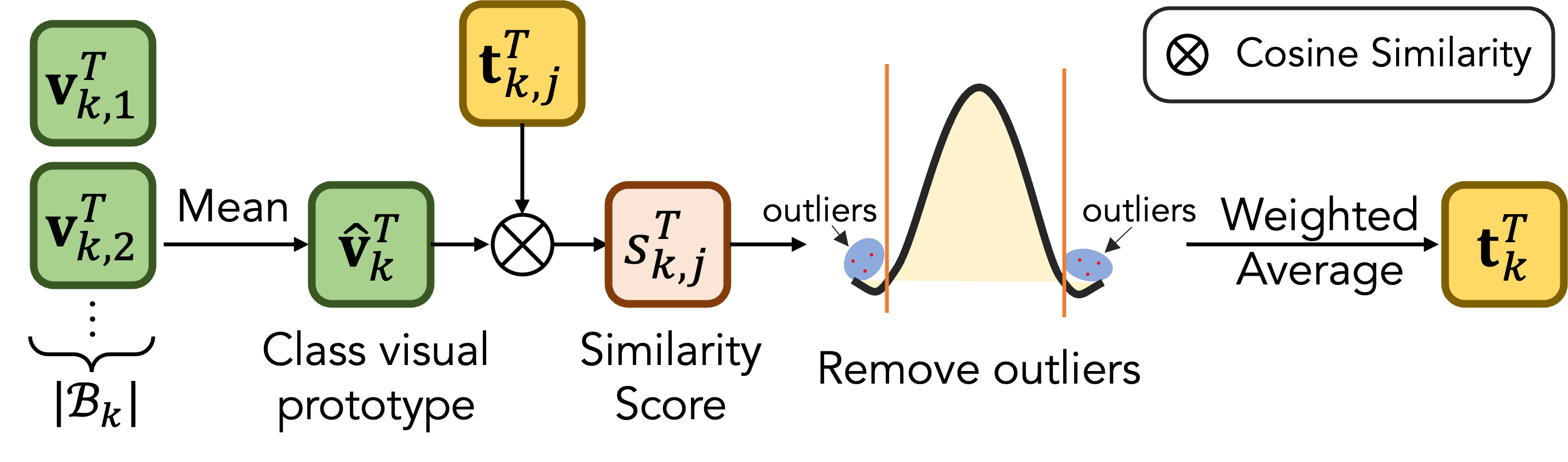}
    \vspace{-16pt}
    \caption{\textbf{Selective Prototype Aggregation.}
For each class $k$, teacher image embeddings $\{\mathbf v^{T}_{k,i}\}_{i\in\mathcal B_k}$ are averaged to form the visual prototype ${\widehat{\mathbf v}}^{T}_{k}$.
Each LLM-generated description is encoded to a teacher text embedding $\mathbf t_{k,j}$ and scored by cosine similarity $s_{k,j}=({\widehat{\mathbf v}}^{T}_{k})^{\top}\mathbf t^{T}_{k,j}$.
A median/MAD threshold $\pm\zeta_s$ removes outlier candidates.
The remaining embeddings are combined with softmax-normalized weights (over kept $j$) to obtain the class textual prototype ${\mathbf t}^{T*}_{k}$.}

    \label{fig:SRP}
    \vspace{-6pt}
\end{figure}

LLM candidates may include ground-view phrases or off-topic wording. As shown in Fig.~\ref{fig:SRP},
visual prototypes are used to \emph{score} these candidates.
We calculate the similarity between the textual embedding $\mathbf{t}^{T}_{k,j}$ of each candidate description for class $k$ and the corresponding visual prototype:
\begin{equation}
s_{k,j}=({\widehat{\mathbf v}}^T_k)^\top\mathbf t^{T}_{k,j}.
\label{eq:score}
\end{equation}
Next, we remove outlying or generic candidates using a median/MAD-based robust $z$-score (closed form in Appendix~\ref{app:srp_details}, Eq.~\ref{eq:mad}). We denote the kept index set by $\mathcal{J}_k$.
The kept candidates are then aggregated based on the \texttt{RS-Flag} acquired from the last step to obtain the final prototype $\mathbf t^{T*}_k$:
\begin{align}
w_{k,j} &\propto \exp\!\big(\beta s_{k,j}+\gamma\,\texttt{RS-Flag}_{k,j}\big),\quad j\in\mathcal{J}_k,\label{eq:weights}\\
\mathbf t^{T*}_k&= \mathrm{norm}\!\Big(\sum_{j\in\mathcal{J}_k}\frac{w_{k,j}}{\sum_{j'\in\mathcal{J}_k}w_{k,j'}}\,\mathbf t_{k,j}\Big),
\label{eq:proto}
\end{align}
where $\texttt{RS-Flag}_{k,j}\!\in\!\{0,1\}$.

This aggregation process is analogous to a parameter-free cross-attention mechanism.
The class visual prototype (${\widehat{\mathbf v}}^T_k$) acts as the \textit{query}, and the candidate text embeddings ($\mathbf{t}^{T}_{k,j}$) serve as both \textit{keys} and \textit{values}. 
The \texttt{RS-Flag} prior is then added as a \textit{calibration} to the scores before the softmax-like normalization (Eq.~\ref{eq:weights}).
Then, the weights $w_{k,j}$ are normalized over the kept set $\mathcal{J}_k$.
In such a way, we obtain the textual prototype $\mathbf t^{T*}_k$ for class $k$ generated by the teacher model. 
This prototype contains richer semantic information compared with solely the class name, and also corresponds to the visual features in the target adaptation scenario. 

\subsection{Tri-Aspect Alignment}
\label{subsec:losses}
After acquiring the visual and textual embeddings of the teacher model, as illustrated in Fig.~\ref{fig:overview}, we apply distillation training to update the learnable prompts in the student model. 
First, we apply the task supervision $\mathcal{L}_{\mathrm{task}}$ (cross-entropy in this classification task) to ensure that the student model achieves the desired task performance.
Additionally, we apply three complementary alignment losses to ensure comprehensive knowledge transfer:

\begin{itemize}
    \item \textbf{Visual alignment ($\mathcal{L}_{\mathrm{img}}$):} We encourage the student's visual embeddings ${\mathbf v}^{S}_i$ to mimic the teacher's ${\mathbf v}^{T}_i$:
\begin{equation}
\mathcal{L}_{\mathrm{img}}=\tfrac{1}{|\mathcal{B}|}\sum_{i\in\mathcal{B}}\!\left(1-({\mathbf v}^{S}_i)^\top{\mathbf v}^{T}_i\right),
\end{equation}
    where $\mathcal{B}$ is the current mini-batch. By injecting learnable prompts into the vision side, we aim to address the \emph{visual rigidity} of previous methods.
    This objective guides the student to ``see" RS-specific geometric patterns (\eg, oblique views) as the teacher model does.
    \item \textbf{Textual alignment ($\mathcal{L}_{\mathrm{text}}$):} This loss provides the primary supervision signal for the student's learnable text prompts. Its effect is to align the student's text embedding (${\mathbf t}^{S}_k$) with the high-quality, visually-verified teacher prototypes (${\mathbf t}^{T*}_k$):
\begin{equation}
\mathcal{L}_{\mathrm{text}}=\tfrac{1}{C}\sum_{k=1}^{C}\!\left(1-({\mathbf t}^{S}_k)^\top{\mathbf t}^{T*}_k\right).
\end{equation}
    By minimizing the cosine distance, the student can produce semantic representations that mimic the rich, LLM-guided knowledge captured by the teacher, thereby combating \emph{semantic poverty}.
    \item \textbf{Similarity logit alignment ($\mathcal{L}_{\mathrm{logit}}$):} This loss aligns the student's predictive output distribution with the teacher's (Eq.~\ref{eq:kd}). In contrast to $\mathcal{L}_{\mathrm{task}}$, which uses a one-hot vector representing only the single ground-truth class, this objective provides a more informative supervisory signal. The effect is that the student is trained to replicate the teacher's \emph{entire probability distribution} over all classes:
\begin{equation}
\mathcal{L}_{\mathrm{logit}} = \frac{1}{|\mathcal{B}|} \sum_{i \in \mathcal{B}} \tau^2\,\mathrm{KL}\!\left( \sigma(\mathbf{s}^{(T)}_{i,\cdot}/\tau)\ \|\ \sigma(\mathbf{s}^{(S)}_{i,\cdot}/\tau) \right).
\label{eq:kd}
\end{equation}
This distribution contains rich, implicit information about inter-class relationships, capturing subtle similarities that are lost in the one-hot target. 
\end{itemize}




The overall objective is a weighted sum:
\begin{equation}
\mathcal{L}=\mathcal{L}_{\mathrm{task}}+\lambda_{\mathrm{img}}\mathcal{L}_{\mathrm{img}}+\lambda_{\mathrm{text}}\mathcal{L}_{\mathrm{text}}+\lambda_{\mathrm{logit}}\mathcal{L}_{\mathrm{logit}}.
\label{eq:overall_loss}
\end{equation}
The teacher is only used during training. 
We present the complexity analysis in Appendix~\ref{app:complexity} and detailed hyper-parameter settings in Appendix~\ref{app:symbols}.

\section{Experiments}
\label{sec:experiments}

\subsection{Experimental Setup}
We evaluate \method{} on standard RS benchmarks for few-shot classification, base-to-novel generalization, and cross-modal retrieval.

\paragraph{Few-shot learning.}
To assess the performance of \method{} under limited supervision, which is common in remote sensing, we evaluate few-shot classification under different settings of labeled examples per class ($K\!\in\!\{1,2,4,8,16\}$). To ensure a fair comparison, we fix identical per-class samples across all runs for each method and $K$.

\paragraph{Base-to-novel generalization.}
Each dataset is split into disjoint base/novel classes. Models are trained on base classes with 16-shot supervision and evaluated on both sets. We report the harmonic mean (HM) following prompt-learning practice~\cite{zhou2022coop,zhou2022cocoop,khattak2023maple,guo2025mmrl}. To prevent information leakage from novel classes, the teacher builds prototypes only for base classes. Please refer to Appendix~\ref{app:kd-mask} for the masking mechanism.

\paragraph{Cross-modal retrieval.}
To perform cross-modal retrieval, we extract image and text representations. These are $\ell_2$-normalized, and we retrieve the most similar samples using cosine similarity (the normalized dot product). We report retrieval recall at R@1, R@5, and R@10 for both image-to-text (I$\rightarrow$T) and text-to-image (T$\rightarrow$I) directions. We also report the mean recall (mR), the average of all six values.

\paragraph{Datasets.}
For classification (few-shot and base-to-novel), we use AID~\cite{xia2017aid}, RESISC-45~\cite{cheng2017resisc45}, EuroSAT~\cite{helber2019eurosat}, WHU-RS19~\cite{xia2010structural}, PatternNet~\cite{zhou2018patternnet}, and UCMerced~\cite{yang2010ucm}, covering diverse spatial resolutions, class counts (19--45), and scene complexities. For cross-modal retrieval, we use RSITMD~\cite{yuan2022rsitmd} and RSICD~\cite{lu2017rsicd}.

\paragraph{Implementation details.}
\label{sec:Imp_details}
Unless noted, the student backbone is \texttt{GeoRSCLIP(ViT-B/32)}~\cite{zhang2024rs5m} and the frozen teacher is \texttt{GeoRSCLIP(ViT-H/14)}~\cite{zhang2024rs5m}. 
All teacher-side visual prototypes and textual prototypes generated by selective prototype aggregation are prepared offline and used only during training. Specifically, we employ the Gemini 2.5 Flash API for LLM-based domain prompting to generate up to 50 class-specific descriptions per class. For training, we use AdamW with a learning rate of $5{\times}10^{-4}$ and a batch size of 4. Few-shot runs train for 100 epochs; base-to-novel and retrieval for 50 epochs.
For the distillation objective, we set the final weights to $\lambda_{\mathrm{img}}{=}0.5$, $\lambda_{\mathrm{text}}{=}0.5$, and $\lambda_{\mathrm{logit}}{=}1.0$, using a 30\% linear warm-up for the $\lambda_{\mathrm{logit}}$ term to ensure stable convergence. The distillation temperature $\tau$ is fixed at 2.
All hyperparameters were selected on a validation split of the AID dataset and then fixed for all other datasets and tasks. We provide a detailed justification for these choices, including sensitivity analysis, in Appendix~\ref{app:hyperparams}. All experiments were run on a single NVIDIA L4 GPU (24 GB).

\begin{table}[t]
  \centering
  \caption{Few-shot classification against state-of-the-art methods. The table presents the average classification accuracy (\%) over 6 datasets. The best-performing result for each $K$-shot setting is in \textbf{bold}, and the runner-up is \underline{underlined}.}
  \label{tab:fewshot}
  \small
  \resizebox{\linewidth}{!}{%
    \setlength{\tabcolsep}{5pt}

    \begin{tabular}{lccccc}
      \toprule
      \textbf{Method} & \textbf{1-shot} & \textbf{2-shot} & \textbf{4-shot} & \textbf{8-shot} & \textbf{16-shot} \\
      \midrule
      
      \rowcolor{TableGray}
      \multicolumn{6}{c}{\textbf{Zero-shot Methods}} \\
      CLIP~\cite{radford2021clip}       & \multicolumn{5}{c}{$63.24$} \\
      RemoteCLIP~\cite{liu2023remoteclip}   & \multicolumn{5}{c}{\underline{$72.92$}} \\
      GeoRSCLIP~\cite{zhang2024rs5m}       & \multicolumn{5}{c}{$\mathbf{72.95}$} \\
      \midrule
      
      \rowcolor{TableGray}
      \multicolumn{6}{c}{\textbf{PEFT Methods}} \\
      CoOp~\cite{zhou2022coop}       & $69.98$ & $78.95$ & $84.52$ & $87.57$ & $90.24$ \\
      CoCoOp~\cite{zhou2022cocoop}   & $70.27$ & $80.56$ & $85.74$ & $88.93$ & $91.41$ \\
      MMRL~\cite{guo2025mmrl}         & $70.57$ & $79.47$ & $84.66$ & $87.48$ & $90.45$ \\
      MaPLe~\cite{khattak2023maple}   & $65.60$ & $76.69$ & $82.48$ & $88.33$ & $90.73$ \\
      PromptKD~\cite{li2024promptkd}  & $69.91$ & $79.64$ & $85.02$ & $88.27$ & $90.86$ \\
      APPLeNet~\cite{jha2023applenet} & \underline{$74.27$} & \underline{$81.79$} & \underline{$86.40$} & \underline{$89.20$} & \underline{$91.61$} \\
      \rowcolor{HighlightPurple}
      \textbf{\method{} (Ours)}                 & $\mathbf{74.27}$ & $\mathbf{81.86}$ & $\mathbf{88.31}$ & $\mathbf{91.85}$ & $\mathbf{93.69}$ \\
      \bottomrule
      
    \end{tabular}%
  }  
    \vspace{-4pt}
\end{table}

\subsection{Few-shot Evaluation}
We compare \method{} with two groups of baselines. (1) Zero-shot VLMs: CLIP~\cite{radford2021clip}, RemoteCLIP~\cite{liu2023remoteclip}, and GeoRSCLIP~\cite{zhang2024rs5m}, all evaluated in the zero-shot mode. (2) Parameter-efficient fine-tuning (PEFT): (a) prompt-learning methods CoOp~\cite{zhou2022coop}, CoCoOp~\cite{zhou2022cocoop}, MaPLe~\cite{khattak2023maple}, PromptKD~\cite{li2024promptkd}, and APPLeNet~\cite{jha2023applenet}; (b) a representation-token based method MMRL~\cite{guo2025mmrl}. To ensure fairness, all PEFT baselines are implemented on the same CLIP-like backbone GeoRSCLIP, keeping the image/text encoders frozen while training only method-specific lightweight parameters.

As shown in Tab.~\ref{tab:fewshot}, \method{} matches or exceeds other methods across all shot levels. At 1-shot, it ties the top method (74.27\%), and its performance margin over the runner-up progressively widens with more training samples, peaking at +2.65\,pp at 8-shot. We attribute this high data efficiency to learnable visual prompts and semantically rich textual prototypes generated by the teacher. 
The textual prototypes are then effectively incorporated via our tri-aspect alignment objective, capturing more supervision signals.
Even if the adopted initialization (GeoRSCLIP) already achieves strong zero-shot performance~\cite{zhang2024rs5m}, with RS-specific optimizations in both teacher and student models, our method yields even more significant improvements.


\begin{table}[t]
  \centering
  \caption{Average base-to-novel results (\%) across all 6 datasets (ViT-B/32 student). We report Base, Novel, and HM.}
  \small
    \vspace{-4pt}
  \label{tab:base2novel_avg}  
  \begin{tabular}{lccc}
    \toprule
    \textbf{Method} & \textbf{Base} & \textbf{Novel} & \textbf{HM} \\
    \midrule
    GeoRSCLIP~\cite{zhang2024rs5m} & $78.10$ & \underline{$79.75$} & $78.81$ \\
    CoOp~\cite{zhou2022coop}       & $91.17$ & $69.52$ & $78.88$ \\
    CoCoOp~\cite{zhou2022cocoop}   & $91.85$ & $70.69$ & $80.06$ \\
    MMRL~\cite{guo2025mmrl}         & $93.75$ & $73.69$ & $82.76$ \\
    MaPLe~\cite{khattak2023maple}  & $92.87$ & $74.66$ & $82.90$ \\
    PromptKD~\cite{li2024promptkd} & $93.67$ & $76.58$ & \underline{$84.20$} \\
    APPLeNet~\cite{jha2023applenet}& \underline{$93.84$} & $75.75$ & $83.84$ \\
    \rowcolor{HighlightPurple}
    \textbf{\method{} (Ours)}                 & $\mathbf{95.64}$ & $\mathbf{79.94}$ & $\mathbf{87.05}$ \\
    \bottomrule
  \end{tabular}
    \vspace{-4pt}
\end{table}

\begin{figure}[t]
    \centering
    \includegraphics[width=1\linewidth]{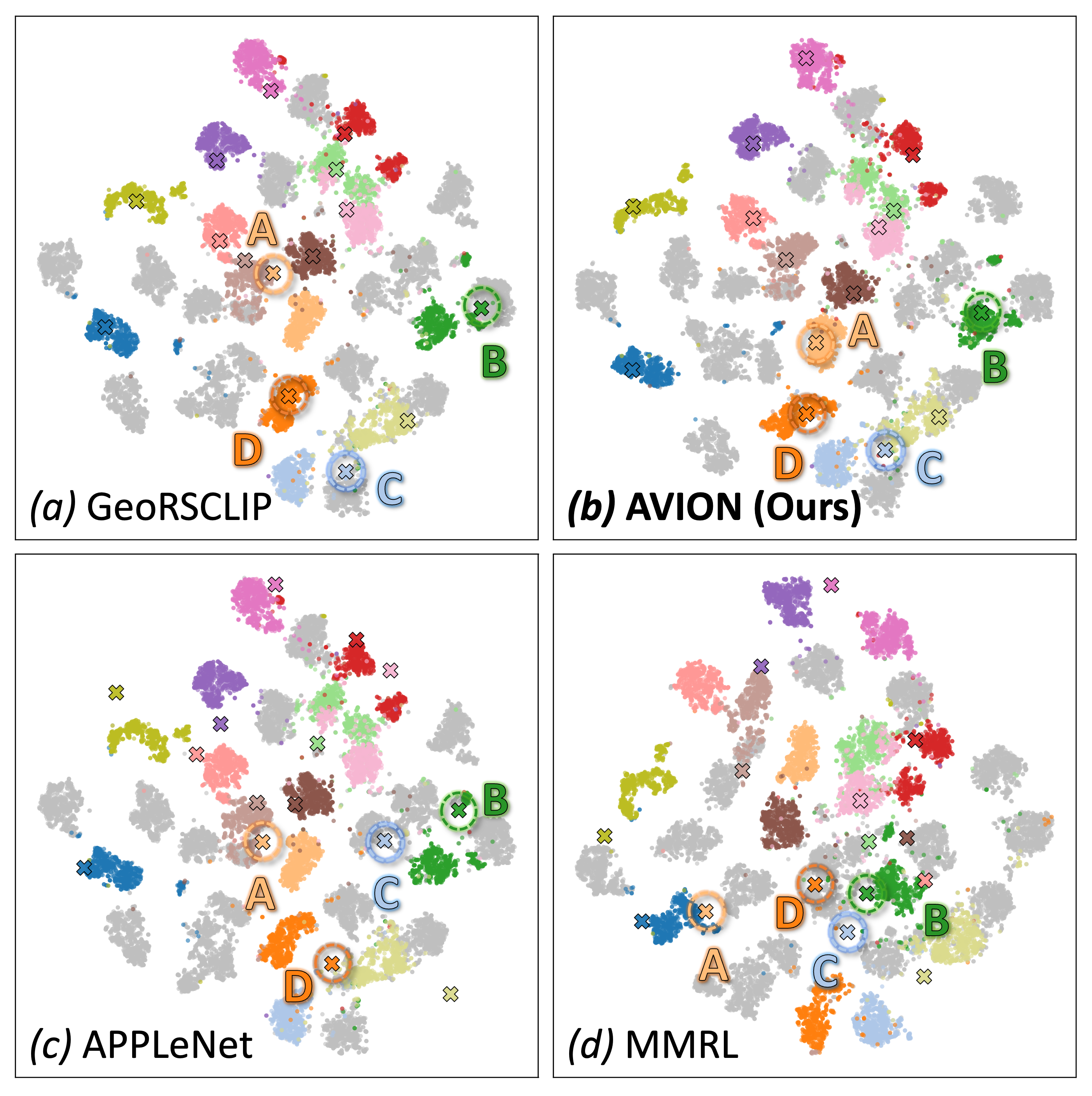}
      \caption{\textbf{t-SNE of base-to-novel visual and text embeddings on RESISC-45.} (a) GeoRSCLIP~\cite{zhang2024rs5m} (b) \textbf{\method{} (ours)} (c) APPLeNet~\cite{jha2023applenet} (d) MMRL~\cite{guo2025mmrl}. Gray dots represent the visual embedding of \textit{base classes}. Colored \textit{dots} represent the \textit{visual embedding of novel classes} with each color standing for a class. Colored \textit{crosses} represent the \textit{text embedding of novel classes} for their class names. A, B, C, and D are four classes.}
    \label{fig:b2ntsne}
    \vspace{-4pt}
\end{figure}

\subsection{Base-to-Novel Generalization}
\label{sec:base2novel}
The base-to-novel setting evaluates the adaptation to base classes and the generalization to novel classes after fine-tuning. 
As shown in Tab.~\ref{tab:base2novel_avg}, with a strong initialization (GeoRSCLIP), all previous methods achieve higher performances on base classes. However, the adaptation also leads to degraded performance on novel classes. In our comparison, \method{} is the only method that surpasses GeoRSCLIP~\cite{zhang2024rs5m} in novel-class performance. Meanwhile, \method{} also achieves the best performance on base classes, surpassing the runner-up by +1.80pp. 

We present the t-SNE visualization of visual and text embeddings for both base and novel classes in Fig.~\ref{fig:b2ntsne}. Gray clusters correspond to base classes, while colored clusters denote novel classes (each color indicates a different class). Colored crosses represent the text embedding of class names for the corresponding colored classes. A, B, C, and D are four representative classes. We highlight their text embeddings with dashed circles. 
Overall, \method{} shows slightly better (\eg, classes A and B) or on-par multimodal alignment on novel classes compared with GeoRSCLIP~\cite{zhang2024rs5m}. It corresponds to the quantitative results in Tab.~\ref{tab:base2novel_avg}. 
By contrast, the other two methods, APPLeNet~\cite{jha2023applenet} and MMRL~\cite{guo2025mmrl}, illustrate misalignment between text and visual embeddings. APPLeNet freezes the visual encoder and updates only the text encoder. While the visual embedding distribution remains consistent with GeoRSCLIP, the text embeddings on these unseen classes shift after tuning. MMRL opens both encoders for training, but without semantically rich text prototypes as guidance, the multimodal gap becomes even larger for unseen classes. 

\begin{table*}[t]
  \centering
  \caption{Cross-modal retrieval (\textbf{R@K}, \%) on RSITMD~\cite{yuan2022rsitmd} and RSICD~\cite{lu2017rsicd} (ViT-B/32 student; protocol details in Sec.~\ref{sec:Imp_details}). \textbf{mR} is the macro-average over $\{\text{T}\rightarrow\text{I},\text{I}\rightarrow\text{T}\}\times\{1,5,10\}$. $\Delta$ is relative to GeoRSCLIP. Our method (\method{}) is lightweight, updating only prompts/heads; see Appendix~\ref{app:complexity} for a full efficiency analysis.}
  \label{tab:retrieval}
  \small
  \resizebox{\textwidth}{!}{
  \begin{tabular}{lllccccccc}
    \toprule
    \multirow{3}{*}{\makecell{\textbf{Testing}\\\textbf{Dataset}}} &
    \multirow{3}{*}{\makecell{\textbf{Training}\\\textbf{Dataset}}} &
    \multirow{3}{*}{\textbf{Method}} &
    \multicolumn{3}{c}{\textbf{Image $\to$ Text}} &
    \multicolumn{3}{c}{\textbf{Text $\to$ Image}} &
    \multirow{2}{*}{\textbf{mR}}\\
    \cmidrule(lr){4-6}\cmidrule(lr){7-9} 
    & & & \textbf{R@1} & \textbf{R@5} & \textbf{R@10} & \textbf{R@1} & \textbf{R@5} & \textbf{R@10} & \\
    \midrule

    \multirow{5}{*}{\textbf{RSITMD}~\cite{yuan2022rsitmd}} &
      RSITMD & ML Retrieval~\cite{al2022multilanguage} &
      $19.09$ & $40.26$ & $54.42$ & $17.61$ & $49.73$ & $60.59$ & $41.38$ \\
    & RET\text{-}3 & CLIP~\cite{radford2021clip} &
      $23.67$ & $50.00$ & $63.27$ & $22.61$ & $55.27$ & $69.87$ & $47.63$ \\
    & RET\text{-}3 + DET\text{-}10 + SEG\text{-}4 & RemoteCLIP~\cite{liu2023remoteclip} &
      $27.88$ & $50.66$ & $65.71$ & $22.17$ & $56.46$ & $67.41$ & $49.38$ \\
    & RSITMD + RSICD & GeoRSCLIP-FT~\cite{zhang2024rs5m} &
      $32.30$ & $53.32$ & $67.92$ & $25.04$ & $57.88$ & $74.38$ & $51.81$ \\
      \rowcolor{HighlightPurple}
    & RSITMD + RSICD & \textbf{\method{} (Ours)} &
      $\mathbf{33.17}$ & $\mathbf{54.46}$ & $\mathbf{69.18}$ & $\mathbf{26.09}$ & $\mathbf{59.22}$ & $\mathbf{75.41}$ & $\mathbf{52.92}$ \\
    \addlinespace[-1pt]

    \midrule

    \multirow{5}{*}{\textbf{RSICD}~\cite{lu2017rsicd}} &
      RSICD & ML Retrieval~\cite{al2022multilanguage} &
      $10.70$ & $29.64$ & $41.53$ & $9.14$ & $29.98$ & $44.59$ & $27.43$ \\
    & RET\text{-}3 & CLIP~\cite{radford2021clip} &
      $17.94$ & $35.94$ & $51.04$ & $13.89$ & $35.15$ & $50.08$ & $33.98$ \\
    & RET\text{-}3 + DET\text{-}10 + SEG\text{-}4 & RemoteCLIP~\cite{liu2023remoteclip} &
      $17.02$ & $39.97$ & $51.51$ & $13.73$ & $34.71$ & $48.28$ & $35.26$ \\
    & RSITMD + RSICD & GeoRSCLIP-FT~\cite{zhang2024rs5m} &
      $21.13$ & $41.72$ & $55.63$ & $15.59$ & $41.19$ & $57.99$ & $38.87$ \\
      \rowcolor{HighlightPurple}
    & RSITMD + RSICD & \textbf{\method{} (Ours)} &
      $\mathbf{22.01}$ & $\mathbf{42.37}$ & $\mathbf{56.42}$ & $\mathbf{16.24}$ & $\mathbf{42.58}$ & $\mathbf{59.21}$ & $\mathbf{39.80}$ \\
    \addlinespace[-1pt]

    \midrule
    \multicolumn{3}{r}{\(\Delta\) vs.\ GeoRSCLIP-FT (mR)} & 
    \multicolumn{7}{c}{\textbf{RSITMD:} {\textcolor{teal}{+$1.11$}} \quad\quad \textbf{RSICD:} {\textcolor{teal}{+$0.93$}}} \\
    \bottomrule
  \end{tabular}}
\end{table*}

\begin{table*}[t] 
  \centering
  \caption{\textbf{Configurations for our incremental ablation study.} Each configuration builds upon the previous one and culminates in the full system (B7).}
  \label{tab:ablation_config}
  \small
  \setlength{\tabcolsep}{5pt}
  \resizebox{\textwidth}{!}{
  \begin{tabular}{llll}
    \toprule
    ID & Configuration Description & ID & Configuration Description \\
    \midrule
    B0 & Shallow text-only prompt (CoOp-style), $\mathcal{L}_{\mathrm{task}}$ only. & B4 & Replace manual prototypes with LLM-generated ones. \\
    B1 & + Deep prompt tuning. & B5 & + \textit{Selective prototype aggregation} (vision-guided). \\
    B2 & + Representation alignment ($\mathcal{L}_{\mathrm{img}}$). & B6 & + Behavioral alignment ($\mathcal{L}_{\mathrm{logit}}$) in a single-stage schedule. \\
    B3 & + Semantic alignment ($\mathcal{L}_{\mathrm{text}}$) with manual textual prototypes. & \textbf{B7} & \textbf{Final Model}: 30\% warm-up (Sec.~\ref{sec:Imp_details}). \\
    \bottomrule
  \end{tabular}
  }
\end{table*}

The teacher textual prototypes in \method{} integrate rich knowledge from LLMs into the distillation process. Through verification with dataset-specific visual prototypes, \method{} accurately aligns various visual appearances with their corresponding labels and descriptions. 
The external knowledge, in addition to class names, facilitates a comprehensive understanding of base classes, which further promotes the generalization to novel classes.



\subsection{Cross-modal Retrieval}
We evaluate cross-modal retrieval using a deployment-oriented protocol. To simulate scenarios where the gallery is fixed, gallery embeddings are pre-computed and are not re-encoded during adaptation. Concretely, \method{} uses a lightweight prompt learning approach, which requires updating only a minimal set of parameters (well below 1\% of the backbone) and no gallery recomputation. This resource-efficient design is detailed in Appendix~\ref{app:complexity}.

As shown in Tab.~\ref{tab:retrieval}, \method{} attains the best mR across both directions with lightweight adaptation on RSITMD and RSICD. 
Compared with GeoRSCLIP-FT, which opens all the parameters for tuning, \method{} improves mR by +1.11\,pp (RSITMD) and +0.93\,pp (RSICD). This indicates  the computational efficiency of our framework. 
When the trainable parameters are fewer, we introduce additional rich knowledge from LLMs to promote the understanding of these RS classes. The results demonstrate that this semantically richer supervision and our tri-aspect alignment objective are more effective than naive full fine-tuning. 

\subsection{Ablation Study}
\label{sec:ablation}
\paragraph{Incremental ablation analysis.}
We conduct an incremental ablation study to quantify the contribution of each component of \method{}. Starting from a minimal prompt-only baseline, we enable deep prompt tuning, textual prototype enhancement, and the tri-aspect alignment objective (\(\mathcal{L}_{\mathrm{img}}, \mathcal{L}_{\mathrm{text}}, \mathcal{L}_{\mathrm{logit}}\)). Eight configurations (B0–B7) and their results are listed in Tab.~\ref{tab:ablation_config} and Tab.~\ref{tab:cls-results}, respectively. All runs share identical training budgets, datasets, and evaluation protocols. To avoid leakage in the base-to-novel protocol, the offline teacher is used only during training, and its guidance (textual prototypes and distillation targets) is restricted to base classes with masked renormalization. More details are presented in Appendix \ref{app:kd-mask}.

Injecting deep prompts (B1) alone improves the base-class accuracy, but also leads to a -16.01\,pp degradation on novel classes. When gradually adding more training samples, the improvement of deep prompts becomes marginal, indicating unstable effects. 


Adding visual alignment (B2) acts as an effective regularizer, partially reversing the generalization degradation (HM +6.03\,pp). 
Except for the 2-shot case, B2 outperforms B0 across all few-shot configurations. This supports the necessity of introducing the visual alignment $\mathcal{L}_{\mathrm{img}}$. 

Introducing our enhanced textual prototypes at this stage delivers the largest cumulative HM improvement (+10.31 pp from B2 to B5). This trajectory is paralleled in the few-shot domain, which sees cumulative gains from B2 to B5 of +2.31 pp (1-shot) and +1.14 pp (16-shot). These results are consistent with the hypothesis that high-quality, visually grounded semantics are pivotal for both generalization and data efficiency.

Finally, adding similarity logit alignment ($\mathcal{L}_{\mathrm{logit}}$, B6) and an optimized warm-up schedule (B7) yields a further HM gain (+4.00 pp from B5 to B7). This final optimization pushes HM, 1-shot, and 16-shot to their peak values. 
Overall, B7 provides the best overall performance, particularly in generalization (HM) and extreme few-shot scenarios.

\begin{table}[t]
  \centering
  \caption{\textbf{Main ablation results for classification.} We report Base, Novel, and Harmonic Mean (HM) accuracy. Metrics are averaged over six datasets.}
  \label{tab:cls-results}
  \small
  \vspace{-4pt}
  \setlength{\tabcolsep}{3pt}
  \resizebox{\linewidth}{!}{
  \begin{tabular}{lcccccccc}
    \toprule
    \multirow{3}{*}{\textbf{ID}} & \multirow{3}{*}{\textbf{Base}} & \multirow{3}{*}{\textbf{Novel}} & \multirow{3}{*}{\textbf{HM}} & \multicolumn{5}{c}{\textbf{K-shot}} \\
    \cmidrule(lr){5-9}
     &  &  &  & 1 & 2 & 4 & 8 & 16 \\
    \midrule
    B0 & $91.17$& $69.52$& $78.88$& $69.98$& $78.95$& $84.52$& $87.57$& $90.24$\\
    B1 & $93.52$ & $53.51$& $66.71$& $66.95$& $84.68$& $84.61$& $87.34$& $91.25$ \\
    B2 & $89.74$& $61.15$& $72.74$& $70.21$& $82.99$ & $85.31$& $89.65$ & $91.51$ \\
    B3 & $91.21$& $68.63$& $78.34$& $71.98$& $83.72$ & $85.90$& $90.43$ & $92.18$ \\
    B4 & $91.03$& $74.08$& $81.67$& $72.15$& $84.71$ & $87.43$& $91.89$ & $92.14$\\
    B5 & $91.71$& $75.87$& $83.05$& $72.52$& $85.06$ & $87.91$& $92.53$ & $92.65$\\
    B6 & $94.82$ & $79.14$& $86.26$& $73.71$& $\mathbf{85.34}$ & $\mathbf{89.28}$ & $\mathbf{92.95}$ & $93.17$\\
    \textbf{B7} & $\mathbf{95.64}$ & $\mathbf{79.94}$ & $\mathbf{87.05}$ & $\mathbf{74.27}$ & $81.86$ & $88.31$ & $91.85$ & $\mathbf{93.69}$ \\
    \bottomrule
  \end{tabular}
  }
  \vspace{-5pt}
\end{table}

\paragraph{Ablating the teacher text-prototype pipeline.}
We fix the model architecture and training objectives for all settings, varying only prototype sources and aggregation in Tab.~\ref{tab:proto_ablation}. Handcrafted multi-templates (P1) serve as a baseline (79.91 HM). Switching to domain-constrained LLM candidates with simple averaging (P3) already improves this to 81.70 HM. This highlights the importance of the additional knowledge introduced by LLMs. \textit{Selective prototype aggregation} (P6) further boosts the score to 83.05 HM. This +1.35 pp gain (P6 vs. P3) isolates the benefit of our aggregation over a simple mean. The value of its \textit{visual} guidance is also supported by P5 outperforming a text-only selector (P4, 82.05 HM). 
We also demonstrate the result of selecting a random subset from candidates (P5), which results in a score of 80.31 HM. Notably, random selection performs even worse than averaging all unconstrained candidates (P2, 80.85 HM). 
The contrast further supports that P6's gains stem from its vision-guided selection criteria.

\begin{table}[t]
  \centering
  \caption{\textbf{Teacher text-prototype ablation (P0–P6).} 
  All settings use the same objectives, with different prototype sources and aggregation. HM averaged over the same datasets/seeds.}
  \label{tab:proto_ablation}
  
  \resizebox{\columnwidth}{!}{%

    \setlength{\tabcolsep}{4pt} 
    \begin{tabular}{l l l l c}
      \toprule
      
      ID & Prototype Source & Aggregation & \texttt{RS-Flag} & HM (\%) \\
      \midrule

      P0 & Manual (single template) & none & -- & $79.52$ \\
      P1 & Manual (multi-templates) & mean & -- & $79.91$ \\
      P2 & LLM (unconstrained) & mean & no & $80.85$ \\
      P3 & LLM (domain-constrained) & mean & yes & $81.70$ \\
      P4 & LLM (domain-constrained) & text-only selection & yes & $82.05$ \\
      P5 & LLM (random subset) & mean & no & $80.31$ \\
      \textbf{P6} & \textbf{LLM (domain-constrained)} & \textbf{\textit{Selective agg.}} & \textbf{yes} & $\mathbf{83.05}$ \\
      \bottomrule
    \end{tabular}%
  }
\end{table}

\begin{figure}[t]
  \centering
  \includegraphics[width=0.9\linewidth]{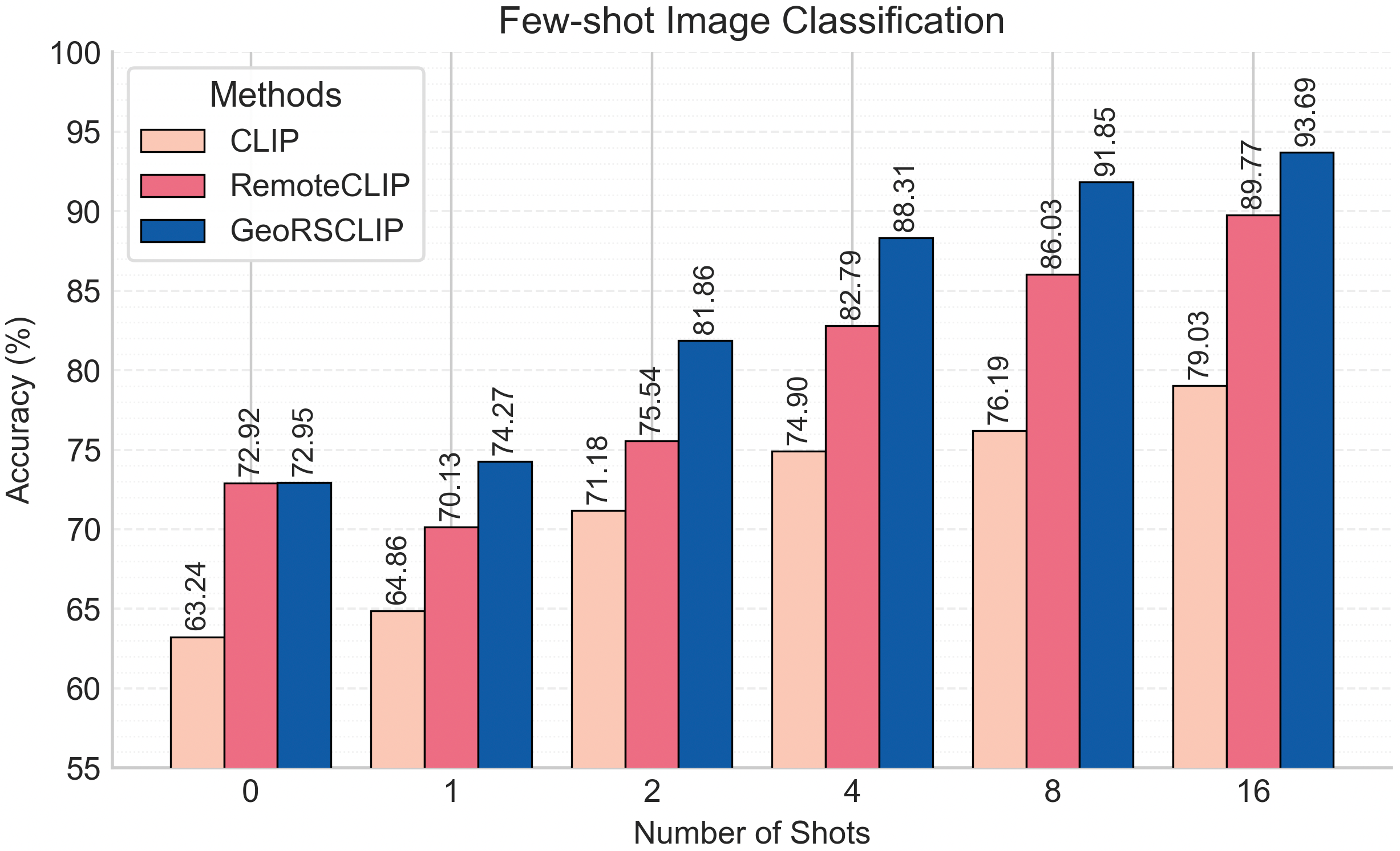}
  \vspace{-8pt}
  \caption{Few-shot classification accuracy of \method{} with different CLIP-based backbones in the few-shot setting.}
  \label{fig:backbone_bar}
  \vspace{-8pt}
\end{figure}

\paragraph{Backbone variants (CLIP family).}
We compare \method{}'s performance using three different ViT-B/32 backbones: CLIP, RemoteCLIP, and GeoRSCLIP. The choice of backbone has a clear impact. As seen in Fig.~\ref{fig:backbone_bar}, the domain-pretrained GeoRSCLIP achieves the highest accuracy across all settings, from zero-shot up to 16-shot. RemoteCLIP, which is also pre-trained for remote sensing, is the clear second-best. The general CLIP backbone performs last in all cases. This result aligns with the expectation that backbones already familiar with aerial imagery provide a stronger starting point for adaptation. We also observe that, starting from 1-shot, the performance for all variants consistently improves over the zero-shot initializations.

\section{Conclusion}
\label{sec:con}
We addressed two challenges in adapting vision–language models to remote sensing imagery: semantic poverty in text representations and visual rigidity in feature adaptation. We proposed \method{}, a knowledge distillation framework in which a frozen teacher constructs semantically rich textual prototypes from LLM-generated descriptions and remote sensing image features, and a lightweight student learns multimodal prompts for both vision and language encoders under a tri-aspect alignment objective. This design allows the student to better accommodate appearance variations in aerial scenes while preserving the alignment structure induced by the teacher. Experiments on six remote sensing benchmarks indicate that \method{} can improve few-shot classification performance without degrading generalization to novel categories.


\section*{Acknowledgments}
This work was supported by Mitacs through the Mitacs Accelerate Program and by TerraSense Analytics (Grant No. IT42711). It was supported in part by the U.S. National Science Foundation (OAC-2118240, HDR Institute: Imageomics).

{
    \small
    \bibliographystyle{ieeenat_fullname}
    \bibliography{main}
}
\clearpage
\setcounter{page}{1}
\maketitlesupplementary
\appendix

\noindent
This appendix is organized as follows:
\begin{itemize}
    \item Section~\ref{app:A} reports additional quantitative results, including
    few-shot classification and base-to-novel generalization across six remote
    sensing benchmarks.

    \item Section~\ref{app:symbols} summarizes the symbols and the default or
    searched hyperparameter ranges used in \method{}.

    \item Section~\ref{sec:appendix_rsdp_examples} presents examples of our
    \textit{LLM-based domain prompting} module for fine-grained remote sensing
    classes, comparing simple templates with richer LLM-generated descriptions.

    \item Section~\ref{app:srp_details} details the selective prototype
    aggregation procedure, including the \texttt{RS-Flag} rules, MAD-based robust
    pruning, and pseudo-code for constructing teacher text prototypes.

    \item Section~\ref{app:kd-mask} describes the masking and renormalization
    strategy for the logit distillation loss $\mathcal{L}_{\mathrm{logit}}$ in
    base-to-novel training.

    \item Section~\ref{app:complexity} analyzes the parameter count, FLOPs
    overhead, runtime, and memory footprint of \method{} on top of a frozen
    GeoRSCLIP backbone.

    \item Section~\ref{app:hyperparams} presents a hyperparameter sensitivity
    analysis and our unified selection protocol shared across datasets and tasks.

    \item Section~\ref{sec:rs_flag_qualitative} provides a qualitative analysis of
    \texttt{RS-Flag}, visualizing how the calibration down-weights RS-agnostic or
    ground-level descriptions to mitigate hallucinations.

    \item Section~\ref{sec:cross_dataset} reports cross-dataset zero-shot transfer
    results to assess generalization and source-domain overfitting.

    \item Section~\ref{sec:llm_ablation} ablates the choice of offline LLM
    generators for domain prompting, demonstrating that \method{} is robust to the
    specific LLM used.

    \item Section~\ref{sec:imagenet} extends the evaluation to broader
    general-domain benchmarks (ImageNet) to study scalability beyond remote
    sensing.
\end{itemize}

\section{Additional Quantitative Results}
\label{app:A}

\begin{figure*}[t]
  \centering
  \subfloat[AID~\cite{xia2017aid}]{\includegraphics[width=0.32\textwidth]{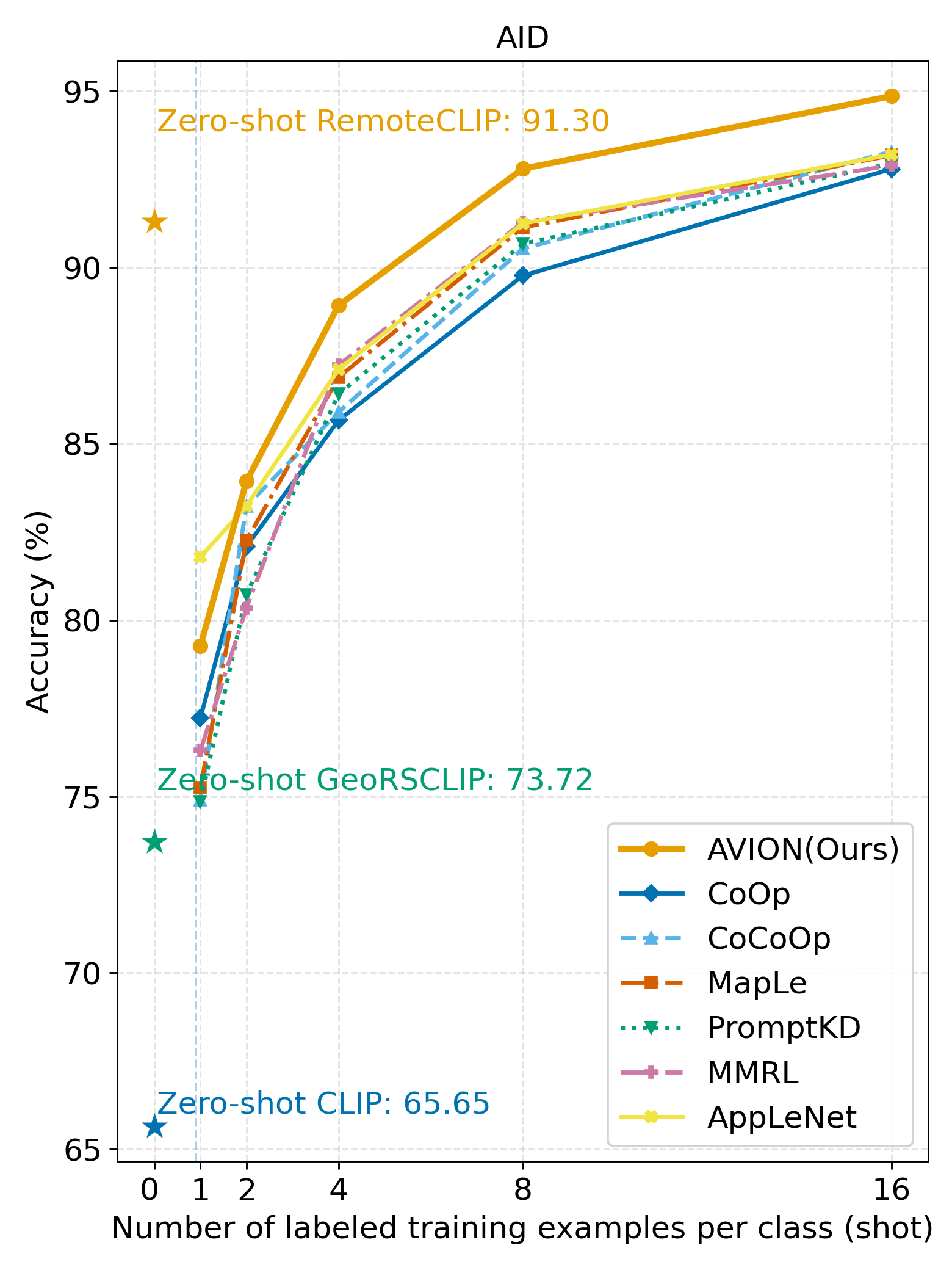}\label{fig:fewshot:aid}}\hfill
  \subfloat[EuroSAT~\cite{helber2019eurosat}]{\includegraphics[width=0.32\textwidth]{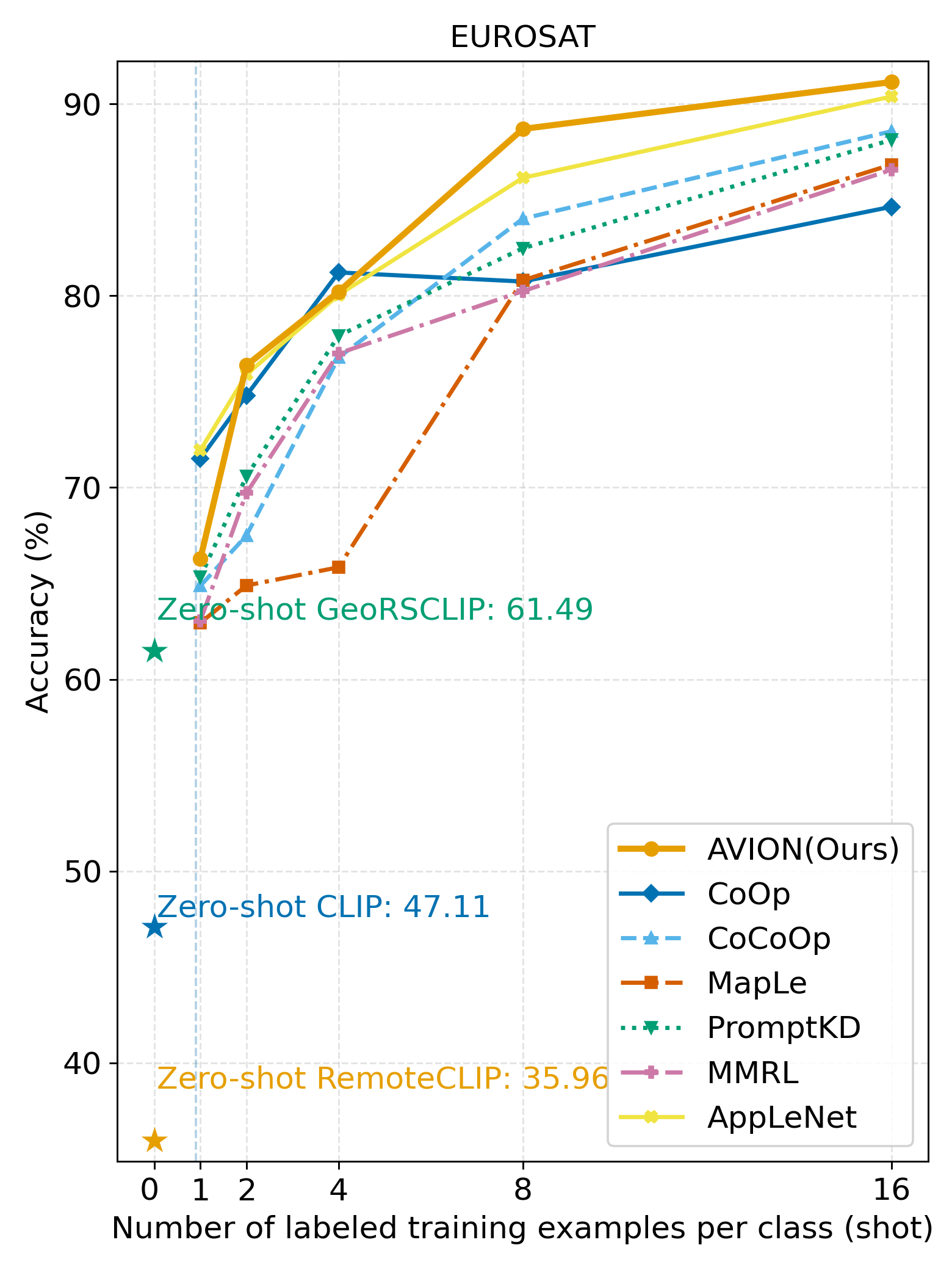}\label{fig:fewshot:eurosat}}\hfill
  \subfloat[RESISC-45~\cite{cheng2017resisc45}]{\includegraphics[width=0.32\textwidth]{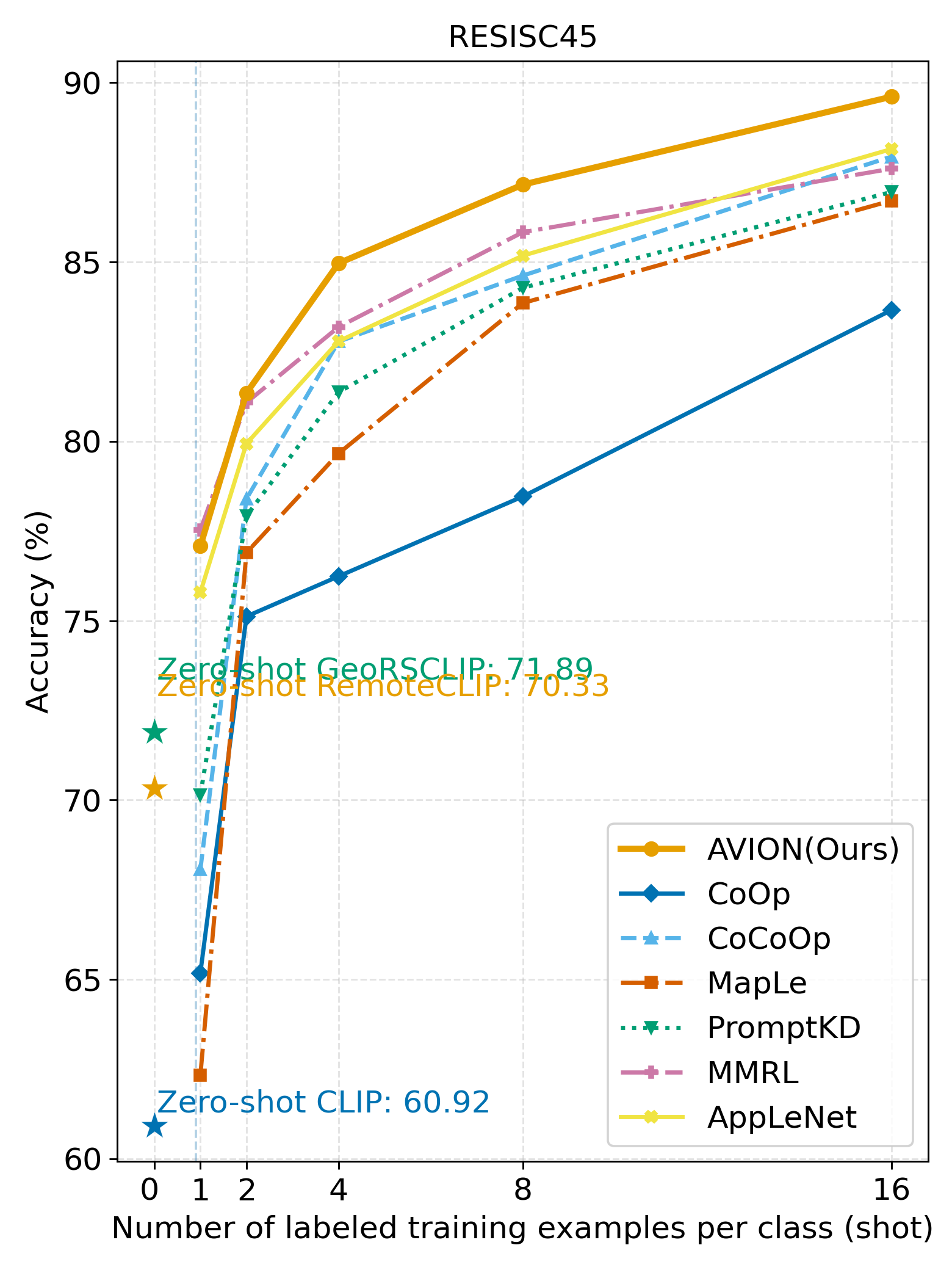}\label{fig:fewshot:resisc45}}\\
  \subfloat[UCMerced~\cite{yang2010ucm}]{\includegraphics[width=0.32\textwidth]{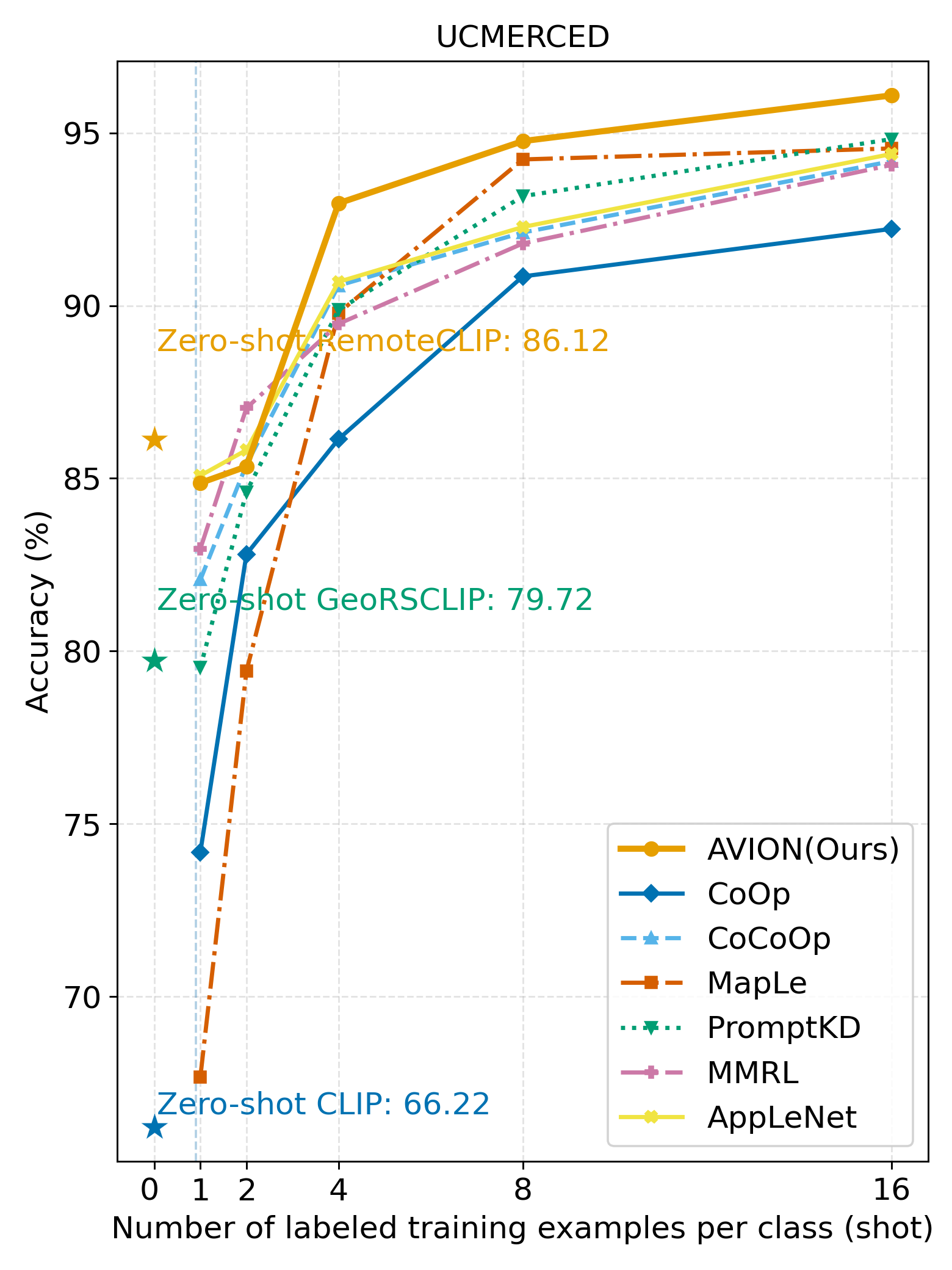}\label{fig:fewshot:ucm}}\hfill
  \subfloat[WHU-RS19~\cite{xia2010structural}]{\includegraphics[width=0.32\textwidth]{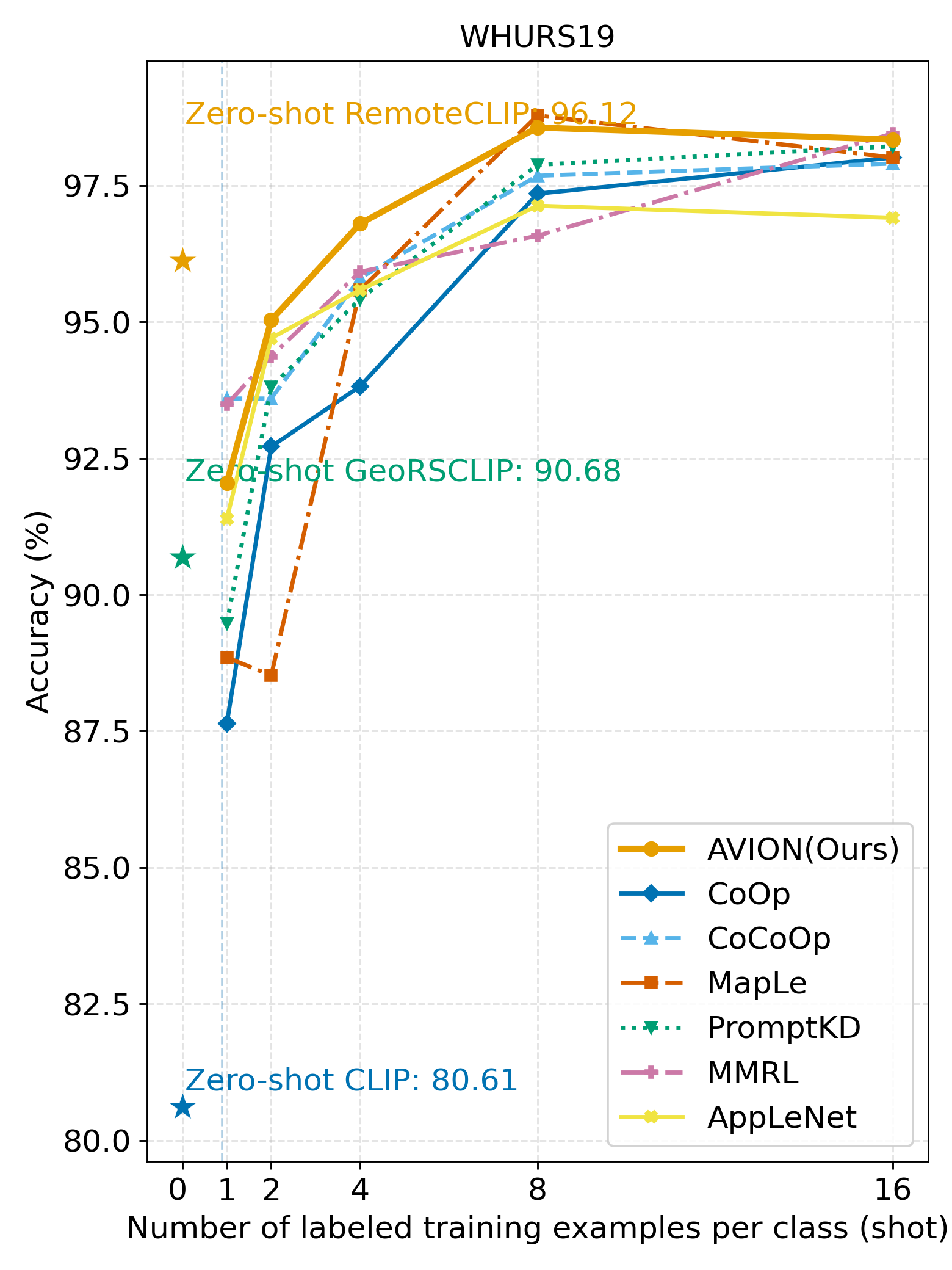}\label{fig:fewshot:whurs19}}\hfill
  \subfloat[PatternNet~\cite{zhou2018patternnet}]{\includegraphics[width=0.32\textwidth]{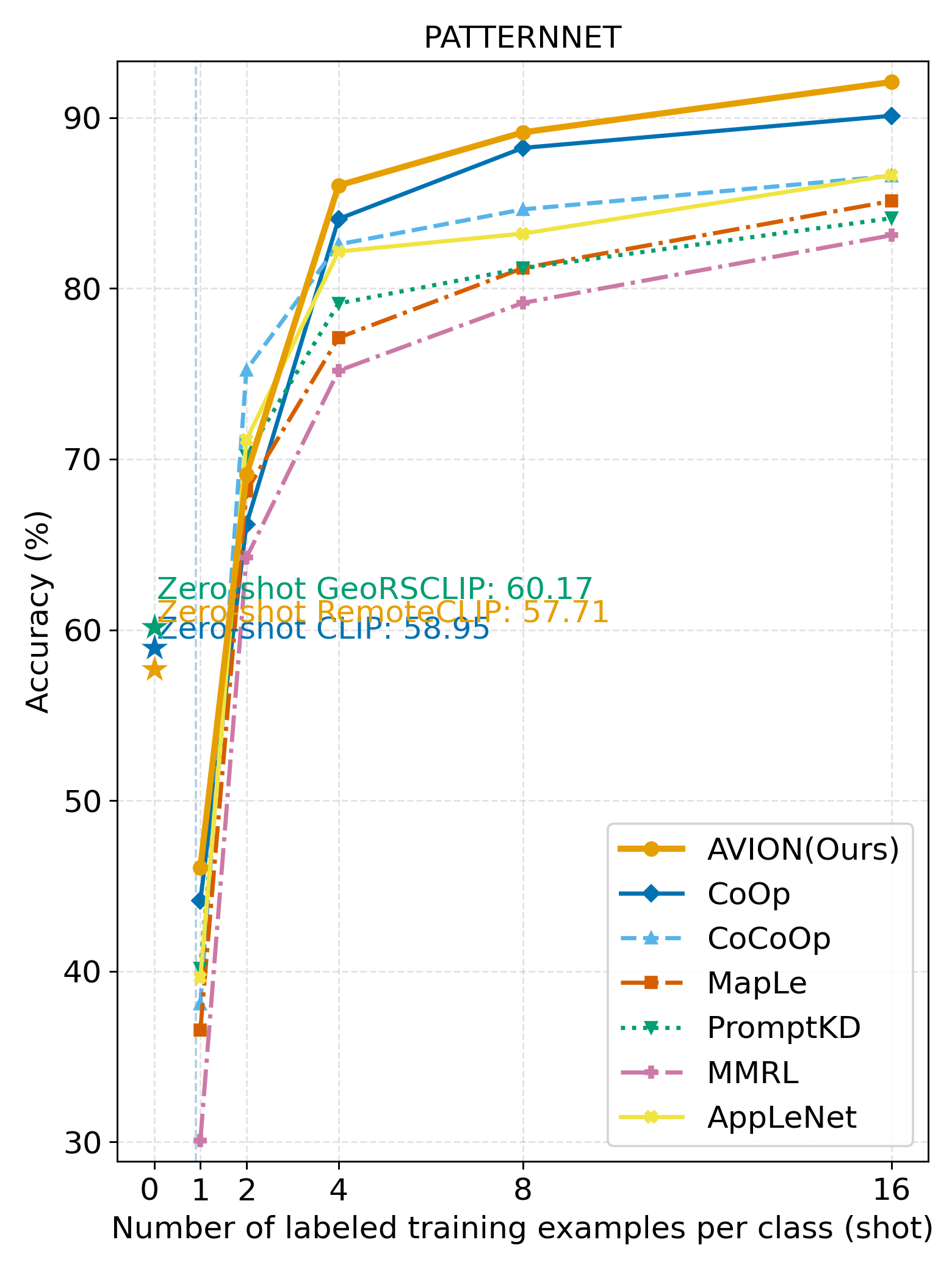}\label{fig:fewshot:patternnet}}
  \caption{Few-shot classification performance (top-1 accuracy, \%) across six remote sensing datasets (AID~\cite{xia2017aid}, EuroSAT~\cite{helber2019eurosat}, RESISC-45~\cite{cheng2017resisc45}, UCMerced~\cite{yang2010ucm}, WHU-RS19~\cite{xia2010structural}, and PatternNet~\cite{zhou2018patternnet}) for different numbers of shots \(K\!\in\!\{1,2,4,8,16\}\).}
  \label{fig:fewshot_grid_datasets}
\end{figure*}

We first provide additional few-shot classification results on six remote sensing datasets.
Fig.~\ref{fig:fewshot_grid_datasets} shows the \(K\)-shot performance of \method{}
and competing methods under \(K\!\in\!\{1,2,4,8,16\}\) across \textbf{AID,
EuroSAT, RESISC-45, UCMerced, WHU-RS19, and PatternNet}.
As \(K\) increases, all methods exhibit smooth performance gains, and \method{}
remains competitive or superior across datasets, with particularly notable
improvements on EuroSAT and UCMerced at higher shot counts.
For completeness, Table~\ref{tab:fewshot_per_dataset_updated}
reports the per-dataset top-1 accuracies for each \(K\).
Overall, \method{} performs on par with or better than strong prompt-learning
and parameter-efficient adaptation baselines across diverse remote sensing scenes.

\begin{figure}[t]
    \centering
    \includegraphics[width=\linewidth]{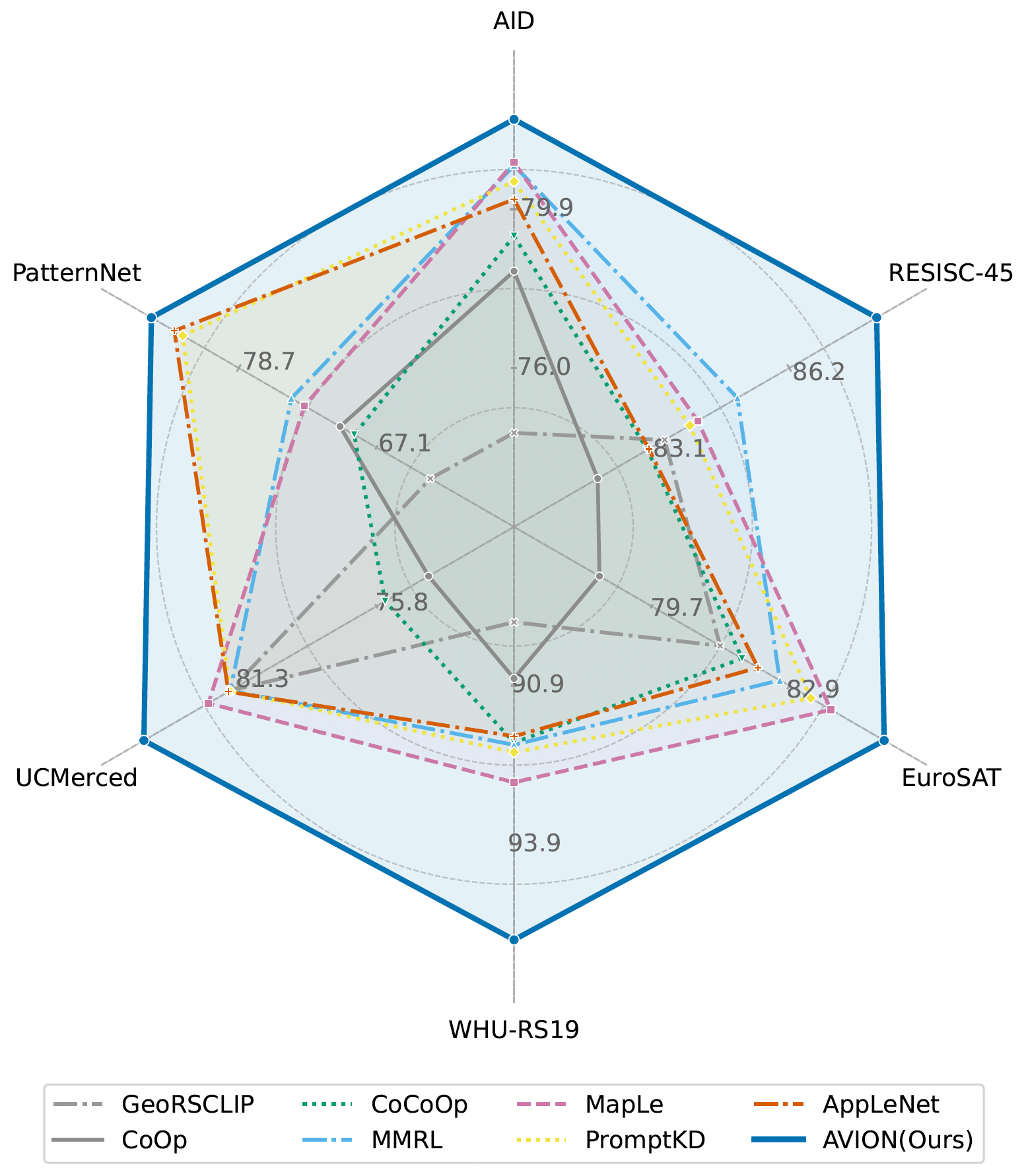}
    \caption{Base-to-novel generalization (HM, \%) across six remote sensing datasets (ViT-B/32 student). The radar chart plots the harmonic mean between base and novel accuracies for \method{} and competing baselines; a larger enclosed area indicates better overall base-to-novel performance.}
    \label{fig:b2n}
\end{figure}

We further complement these results with base-to-novel generalization experiments
on the same six datasets.
Table~\ref{tab:base2novel_detailed} summarizes the detailed Base, Novel, and
harmonic-mean (HM) performance for the ViT-B/32 student model, including the
difference \(\Delta\) between \textbf{Ours} and the strongest competing baseline
for each metric.
Fig.~\ref{fig:b2n} provides a radar-chart view of the HM values across the six
datasets, offering a compact comparison of overall base-to-novel generalization.
\method{} achieves the best HM on all six datasets, improving upon the strongest
baseline by \(+1.08\) to \(+3.16\) points while maintaining competitive Base and
Novel accuracies. On some datasets (e.g., AID, EuroSAT), \method{} trades a small decrease in Novel accuracy for a larger gain on Base and HM, while on the other datasets both Base and Novel improve.

\begin{table}[ht]
  \centering
  \caption{Few-shot classification (top-1, \%) on six remote sensing datasets. 
  Winners are \textbf{bold}; runners-up are \underline{underlined}.}
  \label{tab:fewshot_per_dataset_updated}
  \scriptsize
  \bgroup
  \setlength{\tabcolsep}{3.8pt}
  \renewcommand{\arraystretch}{1.12}
  \begin{tabular}{llccccc}
    \toprule
    \textbf{Dataset} & \textbf{Method} & \textbf{1} & \textbf{2} & \textbf{4} & \textbf{8} & \textbf{16} \\
    \midrule
    \multirow{7}{*}{AID}
      & CoCoOp       & 74.91 & 83.23 & 85.90 & 90.53 & \underline{93.28} \\
      & CoOp         & 77.22 & 82.09 & 85.67 & 89.77 & 92.79 \\
      & MMRL         & 76.30 & 80.35 & \underline{87.23} & \underline{91.28} & 92.89 \\
      & MaPLe        & 75.24 & 82.26 & 86.89 & 91.13 & 93.18 \\
      & PromptKD     & 74.85 & 80.72 & 86.41 & 90.67 & 92.95 \\
      & APPLeNet     & \textbf{81.79} & \underline{83.24} & 87.11 & 91.24 & 93.20 \\
    \rowcolor{HighlightPurple}
      & \textbf{Ours} & \underline{79.27} & \textbf{83.94} & \textbf{88.92} & \textbf{92.80} & \textbf{94.86} \\
    \midrule
    \multirow{7}{*}{EuroSAT}
      & CoCoOp       & 64.87 & 67.49 & 76.78 & 84.01 & 88.57 \\
      & CoOp         & \underline{71.51} & 74.80 & \textbf{81.21} & 80.74 & 84.63 \\
      & MMRL         & 63.05 & 69.72 & 76.99 & 80.23 & 86.56 \\
      & MaPLe        & 62.95 & 64.89 & 65.84 & 80.80 & 86.83 \\
      & PromptKD     & 65.32 & 70.58 & 77.91 & 82.47 & 88.12 \\
      & APPLeNet     & \textbf{71.94} & \underline{75.89} & 80.04 & \underline{86.14} & \underline{90.38} \\
    \rowcolor{HighlightPurple}
      & \textbf{Ours} & 66.29 & \textbf{76.36} & \underline{80.21} & \textbf{88.69} & \textbf{91.14} \\
    \midrule
    \multirow{7}{*}{RESISC-45}
      & CoCoOp       & 68.06 & 78.40 & 82.78 & 84.62 & 87.94 \\
      & CoOp         & 65.18 & 75.12 & 76.24 & 78.46 & 83.66 \\
      & MMRL         & \textbf{77.53} & \underline{81.08} & \underline{83.19} & \underline{85.83} & 87.60 \\
      & MaPLe        & 62.32 & 76.90 & 79.66 & 83.86 & 86.71 \\
      & PromptKD     & 70.14 & 77.92 & 81.37 & 84.28 & 86.95 \\
      & APPLeNet     & 75.79 & 79.94 & 82.79 & 85.18 & \underline{88.15} \\
    \rowcolor{HighlightPurple}
      & \textbf{Ours} & \underline{77.08} & \textbf{81.35} & \textbf{84.96} & \textbf{87.16} & \textbf{89.61} \\
    \midrule
    \multirow{7}{*}{UCMerced}
      & CoCoOp       & 82.07 & 85.40 & 90.58 & 92.12 & 94.18 \\
      & CoOp         & 74.18 & 82.80 & 86.14 & 90.85 & 92.22 \\
      & MMRL         & 82.96 & \textbf{87.04} & 89.47 & 91.80 & 94.07 \\
      & MaPLe        & 67.67 & 79.42 & 89.79 & \underline{94.23} & 94.55 \\
      & PromptKD     & 79.52 & 84.61 & 89.88 & 93.17 & \underline{94.82} \\
      & APPLeNet     & \textbf{85.08} & \underline{85.82} & \underline{90.69} & 92.28 & 94.39 \\
    \rowcolor{HighlightPurple}
      & \textbf{Ours} & \underline{84.87} & 85.34 & \textbf{92.96} & \textbf{94.76} & \textbf{96.09} \\
    \midrule
    \multirow{7}{*}{WHU\textendash RS19}
      & CoCoOp       & \textbf{93.60} & 93.60 & 95.81 & 97.68 & 97.90 \\
      & CoOp         & 87.64 & 92.72 & 93.82 & 97.35 & 98.01 \\
      & MMRL         & \underline{93.49} & 94.37 & \underline{95.92} & 96.58 & \textbf{98.45} \\
      & MaPLe        & 88.85 & 88.52 & 95.59 & \textbf{98.79} & 98.01 \\
      & PromptKD     & 89.47 & 93.81 & 95.42 & 97.88 & 98.22 \\
      & APPLeNet     & 91.39 & \underline{94.70} & 95.59 & 97.13 & 96.91 \\
    \rowcolor{HighlightPurple}
      & \textbf{Ours} & 92.05 & \textbf{95.03} & \textbf{96.80} & \underline{98.56} & \underline{98.34} \\
    \midrule
    \multirow{7}{*}{PatternNet}
      & CoCoOp       & 38.09 & \textbf{75.23} & 82.58 & 84.63 & 86.59 \\
      & CoOp         & \underline{44.13} & 66.18 & \underline{84.07} & \underline{88.23} & \underline{90.11} \\
      & MMRL         & 30.09 & 64.23 & 75.18 & 79.15 & 83.11 \\
      & MaPLe        & 36.57 & 68.14 & 77.11 & 81.19 & 85.12 \\
      & PromptKD     & 40.16 & 70.19 & 79.12 & 81.18 & 84.11 \\
      & APPLeNet     & 39.62 & \underline{71.14} & 82.16 & 83.21 & 86.63 \\
    \rowcolor{HighlightPurple}
      & \textbf{Ours} & \textbf{46.07} & 69.11 & \textbf{86.03} & \textbf{89.14} & \textbf{92.09} \\
    \bottomrule
  \end{tabular}
  \egroup
\end{table}

\begin{table*}[t]
  \centering
  \caption{Detailed base-to-novel results (\%) on six remote sensing datasets (ViT-B/32 student). Baselines:
  GeoRSCLIP~\cite{zhang2024rs5m}, CoOp~\cite{zhou2022coop}, CoCoOp~\cite{zhou2022cocoop},
  MMRL~\cite{guo2025mmrl}, MaPLe~\cite{khattak2023maple}, PromptKD~\cite{li2024promptkd}, and APPLeNet~\cite{jha2023applenet}.
  We report Base, Novel, and HM; \(\Delta\) denotes the difference between \textbf{Ours} and the best-performing baseline for each metric.}
  \label{tab:base2novel_detailed}
  \setlength{\tabcolsep}{6pt}
  \renewcommand{\arraystretch}{1.15}

  \begin{minipage}{0.32\linewidth}
    \centering
    \resizebox{\linewidth}{!}{%
    \begin{tabular}{lccc}
      \toprule
      AID~\cite{xia2017aid} & Base & Novel & HM \\
      \midrule
      GeoRSCLIP~\cite{zhang2024rs5m} & 71.96 & \textbf{76.90} & 74.35 \\
      CoOp~\cite{zhou2022coop}      & \underline{95.53} & 66.41 & 78.35 \\
      CoCoOp~\cite{zhou2022cocoop}   & 94.83 & 68.03 & 79.22 \\
      MMRL~\cite{guo2025mmrl}       & 95.27 & 70.40 & 80.97 \\
      MaPLe~\cite{khattak2023maple}  & 95.10 & 70.60 & \underline{81.04} \\
      PromptKD~\cite{li2024promptkd} & 94.90 & 70.00 & 80.57 \\
      APPLeNet~\cite{jha2023applenet}& 94.60 & 69.50 & 80.13 \\
      \textbf{Ours}                 & \textbf{95.80} & \underline{71.85} & \textbf{82.11} \\
      \midrule
      $\Delta$  & {\color{teal}+0.27} & {\color{teal}-5.05} & {\color{teal}+1.08} \\
      \bottomrule
    \end{tabular}}
    \vspace{3pt}

    \footnotesize (a) AID.
  \end{minipage}
  \hfill
  \begin{minipage}{0.32\linewidth}
    \centering
    \resizebox{\linewidth}{!}{%
    \begin{tabular}{lccc}
      \toprule
      RESISC-45~\cite{cheng2017resisc45} & Base & Novel & HM \\
      \midrule
      GeoRSCLIP~\cite{zhang2024rs5m} & 85.76 & \underline{81.17} & 83.40 \\
      CoOp~\cite{zhou2022coop}      & 90.19 & 74.97 & 81.88 \\
      CoCoOp~\cite{zhou2022cocoop}   & 92.97 & 74.97 & 83.01 \\
      MMRL~\cite{guo2025mmrl}       & \textbf{94.10} & 77.58 & \underline{85.05} \\
      MaPLe~\cite{khattak2023maple}  & 93.50 & 76.50 & 84.15 \\
      PromptKD~\cite{li2024promptkd} & 93.80 & 76.00 & 83.97 \\
      APPLeNet~\cite{jha2023applenet}& 93.00 & 75.00 & 83.04 \\
      \textbf{Ours}                 & \underline{93.90} & \textbf{83.17} & \textbf{88.21} \\
      \midrule
      $\Delta$  & {\color{teal}-0.20} & {\color{teal}+2.00} & {\color{teal}+3.16} \\
      \bottomrule
    \end{tabular}}
    \vspace{3pt}

    \footnotesize (b) RESISC-45.
  \end{minipage}
  \hfill
  \begin{minipage}{0.32\linewidth}
    \centering
    \resizebox{\linewidth}{!}{%
    \begin{tabular}{lccc}
      \toprule
      EuroSAT~\cite{helber2019eurosat} & Base & Novel & HM \\
      \midrule
      GeoRSCLIP~\cite{zhang2024rs5m} & 80.60 & \textbf{82.00} & 81.29 \\
      CoOp~\cite{zhou2022coop}      & 91.09 & 68.94 & 78.48 \\
      CoCoOp~\cite{zhou2022cocoop}   & 92.61 & 73.25 & 81.80 \\
      MMRL~\cite{guo2025mmrl}       & \underline{93.07} & \underline{74.37} & 82.68 \\
      MaPLe~\cite{khattak2023maple}  & 92.95 & 76.40 & \underline{83.87} \\
      PromptKD~\cite{li2024promptkd} & 92.70 & 75.80 & 83.40 \\
      APPLeNet~\cite{jha2023applenet}& 92.20 & 74.10 & 82.17 \\
      \textbf{Ours}                 & \textbf{94.90} & 77.15 & \textbf{85.11} \\
      \midrule
      $\Delta$  & {\color{teal}+1.83} & {\color{teal}-4.85} & {\color{teal}+1.24} \\
      \bottomrule
    \end{tabular}}
    \vspace{3pt}

    \footnotesize (c) EuroSAT.
  \end{minipage}

  \vspace{0.45cm}

  \begin{minipage}{0.32\linewidth}
    \centering
    \resizebox{\linewidth}{!}{%
    \begin{tabular}{lccc}
      \toprule
      WHU-RS19~\cite{xia2010structural} & Base & Novel & HM \\
      \midrule
      GeoRSCLIP~\cite{zhang2024rs5m} & 85.62 & \underline{94.18} & 89.70 \\
      CoOp~\cite{zhou2022coop}      & 93.08 & 88.54 & 90.75 \\
      CoCoOp~\cite{zhou2022cocoop}   & 93.08 & 90.84 & 91.95 \\
      MMRL~\cite{guo2025mmrl}       & \underline{93.92} & 90.14 & 91.99 \\
      MaPLe~\cite{khattak2023maple}  & 93.60 & 91.80 & \underline{92.69} \\
      PromptKD~\cite{li2024promptkd} & 93.40 & 90.90 & 92.13 \\
      APPLeNet~\cite{jha2023applenet}& 93.20 & 90.50 & 91.83 \\
      \textbf{Ours}                 & \textbf{96.85} & \textbf{94.45} & \textbf{95.63} \\
      \midrule
      $\Delta$  & {\color{teal}+2.93} & {\color{teal}+0.27} & {\color{teal}+2.94} \\
      \bottomrule
    \end{tabular}}
    \vspace{3pt}

    \footnotesize (d) WHU-RS19.
  \end{minipage}
  \hfill
  \begin{minipage}{0.32\linewidth}
    \centering
    \resizebox{\linewidth}{!}{%
    \begin{tabular}{lccc}
      \toprule
      UCMerced~\cite{yang2010ucm} & Base & Novel & HM \\
      \midrule
      GeoRSCLIP~\cite{zhang2024rs5m} & 87.27 & \textbf{76.67} & 81.63 \\
      CoOp~\cite{zhou2022coop}      & 94.88 & 60.22 & 73.68 \\
      CoCoOp~\cite{zhou2022cocoop}   & 93.69 & 63.11 & 75.42 \\
      MMRL~\cite{guo2025mmrl}       & \underline{96.58} & 70.67 & 81.62 \\
      MaPLe~\cite{khattak2023maple}  & 96.20 & 72.20 & \underline{82.49} \\
      PromptKD~\cite{li2024promptkd} & 96.00 & 71.00 & 81.63 \\
      APPLeNet~\cite{jha2023applenet}& 95.80 & 71.20 & 81.69 \\
      \textbf{Ours}                 & \textbf{97.90} & \underline{75.21} & \textbf{85.07} \\
      \midrule
      $\Delta$  & {\color{teal}+1.32} & {\color{teal}-1.46} & {\color{teal}+2.58} \\
      \bottomrule
    \end{tabular}}
    \vspace{3pt}

    \footnotesize (e) UCMerced.
  \end{minipage}
  \hfill
  \begin{minipage}{0.32\linewidth}
    \centering
    \resizebox{\linewidth}{!}{%
    \begin{tabular}{lccc}
      \toprule
      PatternNet~\cite{zhou2018patternnet} & Base & Novel & HM \\
      \midrule
      GeoRSCLIP~\cite{zhang2024rs5m} & 57.38 & 67.59 & 62.48 \\
      CoOp~\cite{zhou2022coop}      & 82.22 & 58.06 & 70.14 \\
      CoCoOp~\cite{zhou2022cocoop}   & 83.94 & 53.95 & 68.94 \\
      MMRL~\cite{guo2025mmrl}       & 89.56 & 59.00 & 74.28 \\
      MaPLe~\cite{khattak2023maple}  & 85.89 & 60.45 & 73.17 \\
      PromptKD~\cite{li2024promptkd} & 91.21 & \underline{75.80} & 83.50 \\
      APPLeNet~\cite{jha2023applenet}& \underline{94.22} & 74.20 & \underline{84.21} \\
      \textbf{Ours}                 & \textbf{94.50} & \textbf{77.83} & \textbf{86.16} \\
      \midrule
      $\Delta$  & {\color{teal}+0.28} & {\color{teal}+2.03} & {\color{teal}+1.95} \\
      \bottomrule
    \end{tabular}}
    \vspace{3pt}

    \footnotesize (f) PatternNet.
  \end{minipage}

\end{table*}

\section{Symbols and Default Hyperparameters}
\label{app:symbols}

To facilitate clarity and reproducibility, Table~\ref{tab:symbols} summarizes
the symbols used throughout the paper, together with their definitions and the
default values or hyperparameter search ranges adopted in our experiments.
The table covers model components (encoders and prompts), loss-function terms
(weights and temperatures), and data- or training-related parameters.

\begin{table*}[ht]
\centering
\caption{Symbols and default/search ranges used in \method{}.
In all experiments we fix $K_p{=}30$, $\beta{=}10$, $\gamma{=}2$,
$\zeta_s{=}3.0$.}
\label{tab:symbols}
\small
\setlength{\tabcolsep}{6pt}
\begin{tabular}{l p{0.52\linewidth} l}
\toprule
Symbol & Meaning & Default / Search \\
\midrule
$N$ & \# training images & -- \\
$C$ & \# classes & -- \\
$D$ & image/text embedding dimension & inherited from the VLM \\
$f_T, g_T$ & teacher image/text encoders & frozen \\
$f_S, g_S$ & student image/text encoders & backbones frozen \\
$\mathbf v^{T}_i$ & $\ell_2$-normalized teacher image feature for $x_i$ & unit $\ell_2$-norm \\
$\mathbf t^{T*}_k$ & teacher text prototype for class $k$ & unit $\ell_2$-norm \\
$\mathbf v^{S}_i$ & $\ell_2$-normalized student image feature for $x_i$ & unit $\ell_2$-norm \\
$\mathbf t^{S}_k$ & student text feature for class $k$ & unit $\ell_2$-norm \\
$K_p$ & \# captions per class from \textit{LLM-based domain prompting} & $10$--$50$ \\
$\beta,\gamma$ & weighting in \textit{selective prototype aggregation} (Eq.~\ref{eq:weights}) & $\beta \in [5,20]$, $\gamma \in \{0,2\}$ \\
$\zeta_s$ & MAD-based pruning threshold (Eq.~\ref{eq:mad}) & $2.5$--$3.5$ \\
$\varepsilon$ & numerical stability constant (Eq.~\ref{eq:mad}) & $10^{-8}$ \\
$\tau_s,\tau_t$ & student/teacher logit scales & $\tau_t$ via grid search; $\tau_s$ learnable \\
$\tau$ & temperature of $\mathcal{L}_{\mathrm{logit}}$ (Eq.~\ref{eq:kd}) & $\tau{=}2$ \\
$\lambda_{\mathrm{img}},\lambda_{\mathrm{text}},\lambda_{\mathrm{logit}}$ & loss weights & 0.5, 0.5, 1.0 (fixed; see Appx.~\ref{app:hyperparams}) \\
$P^{(v)}$ & \# visual prompt tokens per layer & $4$--$16$ \\
$P^{(t)}$ & \# text prompt tokens per layer & $2$--$8$ \\
$L_v, L_t$ & \# prompt-injected layers (vision/text) & match ViT/Transformer depth \\
\bottomrule
\end{tabular}
\end{table*}

\section{Examples of \textit{LLM-based Domain Prompting}}
\label{sec:appendix_rsdp_examples}

To provide a concrete illustration of our \textit{LLM-based domain prompting}
module, this section presents example textual candidates generated for several
fine-grained remote sensing classes.

As described in Sec.~\ref{subsec:rsdp}, our goal is to mitigate the
\textit{semantic poverty} of simple templates (e.g., ``a photo of [CLASS]'')
commonly used in prior work. Instead of relying on a single generic template,
we prompt an LLM to generate multiple candidates that are both semantically
richer and explicitly aware of the remote sensing imagery domain. Following the
query design illustrated in Fig.~\ref{fig:RSDP}, we use the following generic
prompt template for each class \texttt{[CLASS]}:
\begin{quote}
\textit{``Generate N overhead-view descriptions of \texttt{[CLASS]} from
satellite imagery, highlighting class-specific scene elements while avoiding
any ground-level terms.''}
\end{quote}
This constraint encourages candidates that are appropriate for overhead
perspectives and contain fine-grained, domain-specific details.

Table~\ref{tab:rsdp_examples} compares standard simple templates with the
candidates produced by our \textit{LLM-based domain prompting} process. The
richer descriptions provide a stronger semantic foundation for the teacher
model's prototypes, enabling our framework to better distinguish between
challenging, fine-grained categories such as ``industrial area'' and
``commercial area''.

\begin{table*}[ht]
\centering
\caption{Examples of LLM-generated candidate descriptions from our
\textit{LLM-based domain prompting} module, compared to standard simple
templates. The generated candidates provide richer semantic and visual cues
specific to remote sensing imagery.}
\label{tab:rsdp_examples}
\resizebox{\textwidth}{!}{%
\begin{tabular}{l|l|l}
\toprule
\textbf{Class Name} & \textbf{Simple Template} & \textbf{LLM-Generated Candidates} \\
\midrule
\texttt{airport} & a photo of an airport & \begin{tabular}[t]{@{}l@{}}
- Overhead airport imagery reveals aprons (with parked aircraft at stands) adjacent to terminals, \\
  \quad connected to the airfield via taxilanes and the taxiway system. \\
- An aerial view of an airport showing multiple long, paved runways and taxiways. \\
- Satellite imagery capturing a large airfield complex with terminal buildings and hangars.
\end{tabular} \\
\midrule
\texttt{industrial area} & a photo of an industrial area & \begin{tabular}[t]{@{}l@{}}
- A top-down view of an industrial zone characterized by large factory buildings and smokestacks. \\
- Aerial imagery of a manufacturing plant complex with large warehouses and shipping depots. \\
- A satellite photo showing a dense cluster of large-scale industrial structures and storage tanks.
\end{tabular} \\
\midrule
\texttt{commercial area} & a photo of a commercial area & \begin{tabular}[t]{@{}l@{}}
- An overhead view of a commercial district featuring multi-story office buildings and retail centers. \\
- Nadir-view imagery of a dense downtown business area with high-rise buildings and streets. \\
- An aerial photo of a shopping center or mall surrounded by a large, organized parking lot.
\end{tabular} \\
\midrule
\texttt{forest} & a photo of a forest & \begin{tabular}[t]{@{}l@{}}
- A vast expanse of dense, green forest canopy as seen from a satellite. \\
- An aerial view looking straight down at a dense woodland of coniferous or deciduous trees. \\
- A top-down remote sensing image of a large, contiguous area covered by trees.
\end{tabular} \\
\bottomrule
\end{tabular}%
}
\end{table*}

\section{Details of \textit{Selective Prototype Aggregation}}
\label{app:srp_details}

This section provides a detailed definition of our RS-Flag prior and a
step-by-step pseudo-code for the \textit{selective prototype aggregation}
process introduced in Sec.~\ref{subsec:rsdp}.

\paragraph{RS-Flag rules.}
The RS-Flag indicator $\mathrm{RS\mbox{-}Flag}_{k,j} \in \{0,1\}$ is set to
$1$ if a generated caption meets all of the following criteria, and $0$
otherwise:
\begin{enumerate}\setlength{\itemsep}{0pt}
    \item \textbf{Token constraints:} The caption must contain at least one
    RS-positive token and no RS-negative tokens. We use the following
    case-insensitive, word-boundary–matched lists:
    \begin{itemize}\setlength{\itemsep}{0pt}
        \item \textbf{Positive:} \texttt{\{overhead, aerial view, satellite imagery, nadir, orthorectified, multispectral, SAR\}}.
        \item \textbf{Negative:} \texttt{\{street, indoor, selfie, portrait, close-up, ground level\}}.
    \end{itemize}
    \item \textbf{Length constraints:} The caption length must be between 6 and
    20 words (whitespace-delimited).
    \item \textbf{Content cues (optional):} The caption may optionally include
    class-specific cues (e.g., geometry- or infrastructure-related terms),
    which are captured implicitly by the LLM and the RS-Flag token lists above.
\end{enumerate}

\begin{center}
\fbox{\parbox{0.96\linewidth}{
\textbf{Input:} teacher features $\{\mathbf v^{T}_i\}$, caption embeddings
$\{\mathbf t_{k,j}\}$; hyperparameters $(\zeta_s,\beta,\gamma,\varepsilon)$. \\[2pt]
\textbf{For each class $k$:}\\
1) $\mathcal{B}_k \leftarrow \{ i \mid y_i {=} k \}$; compute scores
$s_{k,j} \leftarrow \frac{1}{|\mathcal{B}_k|} \sum_{i \in \mathcal{B}_k}
(\mathbf v^{T}_i)^\top \mathbf t_{k,j}$.\\
2) $m_k \leftarrow \mathrm{median}(\{s_{k,j}\}_j)$;
$\Delta_{k,j} \leftarrow \lvert s_{k,j} - m_k \rvert$;
$\mathrm{MAD}_k \leftarrow \mathrm{median}(\{\Delta_{k,j}\}_j)$;\\
\phantom{2)} $z_{k,j} \leftarrow \Delta_{k,j} / (\mathrm{MAD}_k + \varepsilon)$.\\
3) $\mathcal{J}_k \leftarrow \{ j \mid z_{k,j} \le \zeta_s \}$ \hfill
\textit{// robust pruning of outlier captions}\\
4) Compute weights $w_{k,j}$ for $j \in \mathcal{J}_k$ using Eq.~\ref{eq:weights},
where the RS-Flag prior is added as a calibration term to the similarity
scores before softmax normalization.\\
5) $\mathbf t^{T*}_k \leftarrow
\dfrac{\sum_{j \in \mathcal{J}_k} w_{k,j} \mathbf t_{k,j}}%
{\left\lVert \sum_{j \in \mathcal{J}_k} w_{k,j} \mathbf t_{k,j} \right\rVert_2}$
\hfill \textit{// $\ell_2$-normalization}\\[2pt]
\textbf{Output:} $\{\mathbf t^{T*}_k\}_{k=1}^{C}$.
}}
\end{center}

\paragraph{Selective prototype aggregation and robust pruning.}
The \textit{selective prototype aggregation} algorithm (shown in the box above) is performed offline during training. Its core steps involve scoring
caption candidates (Step 1), performing robust pruning (Steps 2-3), and then
aggregating the remaining candidates using the RS-Flag prior (Steps 4-5) to
obtain the final teacher text prototype for each class.

The robust $z$-score used in Step 2 is computed using the median absolute
deviation (MAD). Let
$m_k = \mathrm{median}(\{s_{k,j}\}_j)$,
$\Delta_{k,j} = \lvert s_{k,j} - m_k \rvert$, and
$\mathrm{MAD}_k = \mathrm{median}(\{\Delta_{k,j}\}_j)$.
The $z$-score is defined as
\begin{equation}
z_{k,j} = \frac{\Delta_{k,j}}{\mathrm{MAD}_k + \varepsilon}\,,
\label{eq:mad}
\end{equation}
which downweights or removes outlier captions whose scores deviate strongly
from the median.

\section{Masking and Renormalization for \texorpdfstring{$\mathcal{L}_{\mathrm{logit}}$}{L\_logit} in Base to Novel Training}
\label{app:kd-mask}

Let $\mathcal{Y}_{\mathrm{base}} \subset \mathcal{Y}$ denote the set of base
classes and $\mathcal{Y}_{\mathrm{novel}} = \mathcal{Y} \setminus
\mathcal{Y}_{\mathrm{base}}$ the set of novel classes. During base-to-novel
training, we restrict the distillation distributions to base classes by
masking scores for $\mathcal{Y}_{\mathrm{novel}}$ and renormalizing over
$\mathcal{Y}_{\mathrm{base}}$:
\begin{align}
\tilde{q}^{(T)}_{i,k} &=
\begin{cases}
\displaystyle \frac{\exp\!\big(s^{(T)}_{i,k} / \tau\big)}
{\sum_{k' \in \mathcal{Y}_{\mathrm{base}}} \exp\!\big(s^{(T)}_{i,k'} / \tau\big)}
& k \in \mathcal{Y}_{\mathrm{base}}, \\[6pt]
0 & k \in \mathcal{Y}_{\mathrm{novel}}\,,
\end{cases} \\
\tilde{p}^{(S)}_{i,k} &=
\begin{cases}
\displaystyle \frac{\exp\!\big(s^{(S)}_{i,k} / \tau\big)}
{\sum_{k' \in \mathcal{Y}_{\mathrm{base}}} \exp\!\big(s^{(S)}_{i,k'} / \tau\big)}
& k \in \mathcal{Y}_{\mathrm{base}}, \\[6pt]
0 & k \in \mathcal{Y}_{\mathrm{novel}}\,.
\end{cases}
\end{align}
Here $s^{(T)}_{i,k}$ and $s^{(S)}_{i,k}$ are the teacher and student logits
for image $x_i$ and class $k$, and $\tau$ is the distillation temperature
used in $\mathcal{L}_{\mathrm{logit}}$ (Eq.~\ref{eq:kd}). The
$\mathcal{L}_{\mathrm{logit}}$ term then becomes
\[
\mathcal{L}_{\mathrm{logit}}(x_i)
= \tau^2 \, \mathrm{KL}\!\left(
\tilde{q}^{(T)}_{i,\cdot} \,\middle\|\, \tilde{p}^{(S)}_{i,\cdot}
\right),
\]
which ensures that novel classes neither appear in the denominator nor
contribute to the divergence during training.

\section{Complexity and Inference Overhead}
\label{app:complexity}

We quantify the additional parameters and computational overhead introduced by AVION
on top of a frozen GeoRSCLIP backbone. Unless otherwise stated, all numbers below
correspond to our default ViT-B/32 student with deep prompts in both encoders.

\paragraph{Parameter count.}
For vision and text branches with widths $D_v$ and $D_t$ and $L_v,L_t$ transformer
layers that receive prompts, the total number of prompt parameters is
\[
\#\theta_{\text{prompts}}
\;=\;
L_v\, P^{(v)}\, D_v \;+\; L_t\, P^{(t)}\, D_t,
\]
where $P^{(v)}$ and $P^{(t)}$ denote the number of learnable prompt tokens per
layer on the vision and text sides, respectively (see Tab.~\ref{tab:symbols}).
In our main configuration (GeoRSCLIP ViT-B/32 student), the vision encoder has
$D_v{=}768$, $L_v{=}12$, $P^{(v)}{=}8$, and the text encoder has $D_t{=}512$,
$L_t{=}12$, $P^{(t)}{=}4$. Then
\[
12{\cdot}8{\cdot}768 \;+\; 12{\cdot}4{\cdot}512 \;=\; 98{,}304
\]
trainable prompt parameters, which is well below $1\%$ of a ViT-B/32 backbone.
All other backbone weights remain frozen; beyond these prompts, only a handful
of scalar hyperparameters (e.g., the learnable logit scale $\tau_s$) are updated.

\paragraph{FLOPs overhead (self-attention and MLP).}
Let $N$ be the per-layer token count and $P$ the number of prompt tokens added
to that layer. The self-attention cost scales from $N^2$ to $(N{+}P)^2$, so the
\emph{relative} attention overhead is
\[
r_{\text{attn}}
\;\approx\;
\frac{(N + P)^2 - N^2}{N^2}
\;=\;
2\frac{P}{N} + \left(\frac{P}{N}\right)^2.
\]
The MLP cost scales approximately linearly with sequence length and therefore
grows proportionally to $P/N$.

\textbf{Examples (our setup).}
For the vision branch, ViT-B/32 at input resolution $224^2$ has $N_v{=}49$ patch
tokens; with $P^{(v)}{=}8$,
\[
r_{\text{attn}}^{(v)} \!\approx\! 2\cdot \tfrac{8}{49}
+ \bigl(\tfrac{8}{49}\bigr)^2 \!\approx\! 35.3\%,
\]
and the per-layer MLP overhead is $\approx P^{(v)}/N_v \!\approx\! 16.3\%$.
(We ignore the class token here for a rough estimate.)
For the text branch, the CLIP text encoder uses $N_t{=}77$ tokens; with
$P^{(t)}{=}4$,
\[
r_{\text{attn}}^{(t)} \!\approx\! 2\cdot \tfrac{4}{77}
+ \bigl(\tfrac{4}{77}\bigr)^2 \!\approx\! 10.7\%,
\]
and the MLP overhead is $\approx P^{(t)}/N_t \!\approx\! 5.2\%$.
Because ViT-B/32 has shorter vision sequences than ViT-B/16, the same $P$
yields a larger \emph{relative} attention overhead; reducing $P$ or using
shallower prompting (smaller $L_v,L_t$) lowers this cost. In practice, these
per-layer increases translate into a moderate end-to-end FLOPs and latency
overhead compared to the frozen backbone.

\paragraph{Runtime and memory.}
Table~\ref{tab:runtime} reports the measured runtime of our ViT-B/32 student
(frozen backbone + deep prompts). We report latency, throughput, and memory
consumption over different batch sizes at $224^2$ resolution. Throughput is
computed as \textit{batch size / latency}; exact numbers may vary across
hardware and software stacks. The teacher is used only during training, so it
incurs zero cost at inference.

\begin{table}[ht]
  \centering
  \caption{Runtime with ViT-B/32 student (frozen backbone + deep prompts).
  Teacher is train-time only and has zero inference cost. Throughput is computed
  as batch/latency; exact numbers may vary by hardware and software stack.}
  \label{tab:runtime}
  \setlength{\tabcolsep}{5pt}
  \renewcommand{\arraystretch}{1.05}
  \begin{adjustbox}{max width=\columnwidth}
  \begin{tabular}{lccccc}
    \toprule
    Setting & Batch & Res. & Lat. (ms) & img/s & Mem (MB) \\
    \midrule
    Student & 64  & $224^2$ & 27.4  & 2334 & 1500 \\
    Student & 256 & $224^2$ & 109.7 & 2334 & 5800 \\
    Teacher (train-time only) & 64 & $224^2$ & n/a & n/a & n/a \\
    \bottomrule
  \end{tabular}
  \end{adjustbox}
\end{table}

\section{Hyperparameter Sensitivity Analysis}
\label{app:hyperparams}

In this section, we provide a detailed justification of the hyperparameter
choices used for our \method{} framework, focusing on (i) our parameter
selection protocol, (ii) the effect of loss weights ($\lambda$) and scheduling,
and (iii) our choice of distillation temperature ($\tau$).

\paragraph{Parameter Selection Protocol and Fixed Settings.}
To ensure robustness and avoid over-tuning, all hyperparameters were selected
\emph{once} on a validation split derived from the \textbf{AID} dataset's base
classes. These exact settings (e.g., learning rates, loss weights, warm-up
schedule) were then frozen and applied to all six classification datasets and
both retrieval datasets across all three experimental protocols (few-shot,
base-to-novel, and retrieval). The strong performance across all tasks suggests
the robustness of this single set of hyperparameters. Furthermore, following
prior work on logit-based distillation~\cite{li2024promptkd}, we use a default
distillation temperature of $\tau{=}2$ and do not tune it as part of our
selection process.

\paragraph{Analysis of Loss Weights and Scheduling.}
We conducted a detailed ablation study, summarized in
Table~\ref{tab:appendix_weights_warmup}, to validate our final loss weight
configuration ($\lambda_{\mathrm{img/text}}{=}0.5$,
$\lambda_{\mathrm{logit}}{=}1.0$) and the 30\% linear warm-up strategy on the
$\mathcal{L}_{\mathrm{logit}}$ term. The results lead to three key conclusions:

\begin{table}[ht]
\centering
\caption{Ablation on loss weights ($\lambda$) and scheduling strategies. We
report the average harmonic mean (HM, \%) for base-to-novel generalization.}
\label{tab:appendix_weights_warmup}
\small
\setlength{\tabcolsep}{6pt}
\begin{tabular}{l ccc c}
\toprule
Setting & $\lambda_{\mathrm{img}}$ & $\lambda_{\mathrm{text}}$ & $\lambda_{\mathrm{logit}}$ & Avg. HM \\
\midrule
Only $\mathcal{L}_{\mathrm{logit}}$ & 0   & 0   & 1.0 & 83.14 \\
All 1.0 (2-stage)                  & 1.0 & 1.0 & 1.0 & 85.47 \\
All 1.0 (Warm-up)                  & 1.0 & 1.0 & 1.0 & 86.91 \\
\textbf{Ours (Warm-up)}            & \textbf{0.5} & \textbf{0.5} & \textbf{1.0} & \textbf{87.05} \\
\bottomrule
\end{tabular}
\end{table}

\begin{enumerate}
    \item \textbf{The tri-aspect alignment losses are critical and complementary.}
    Relying only on $\mathcal{L}_{\mathrm{logit}}$ results in a significantly
    lower HM (83.14\%), indicating that the representation alignment
    $\mathcal{L}_{\mathrm{img}}$ and semantic alignment $\mathcal{L}_{\mathrm{text}}$
    play a critical role in achieving top performance.

    \item \textbf{Our chosen weights ($\lambda_{\mathrm{img/text}}{=}0.5$) provide a better balance.}
    Setting all three weights to 1.0 yields a strong 86.91\% HM, but it is
    slightly outperformed by our final configuration at 87.05\% HM. Comparing
    ``All 1.0 (Warm-up)'' and ``Ours (Warm-up)'' isolates the effect of
    $\lambda_{\mathrm{img}}$ and $\lambda_{\mathrm{text}}$, suggesting that
    giving these embedding-level alignment terms a higher weight (1.0 vs.\ 0.5)
    can slightly over-regularize the student and hinder its optimal adaptation
    to the task.

    \item \textbf{The 30\% warm-up is the superior scheduling strategy.}
    Comparing ``All 1.0 (2-stage)'' (85.47\%) and ``All 1.0 (Warm-up)''
    (86.91\%) isolates the impact of the scheduling policy. The smoother
    30\% warm-up schedule yields a clear HM improvement over the hard 2-stage
    scheme, and our final setting (``Ours (Warm-up)'') further reaches 87.05\%.
    This supports our choice of a warm-up strategy for $\mathcal{L}_{\mathrm{logit}}$
    as a more stable and robust schedule.
\end{enumerate}

\section{Qualitative Analysis of RS-Flag}
\label{sec:rs_flag_qualitative}

\begin{figure}[htbp]
    \centering
    \includegraphics[width=\linewidth]{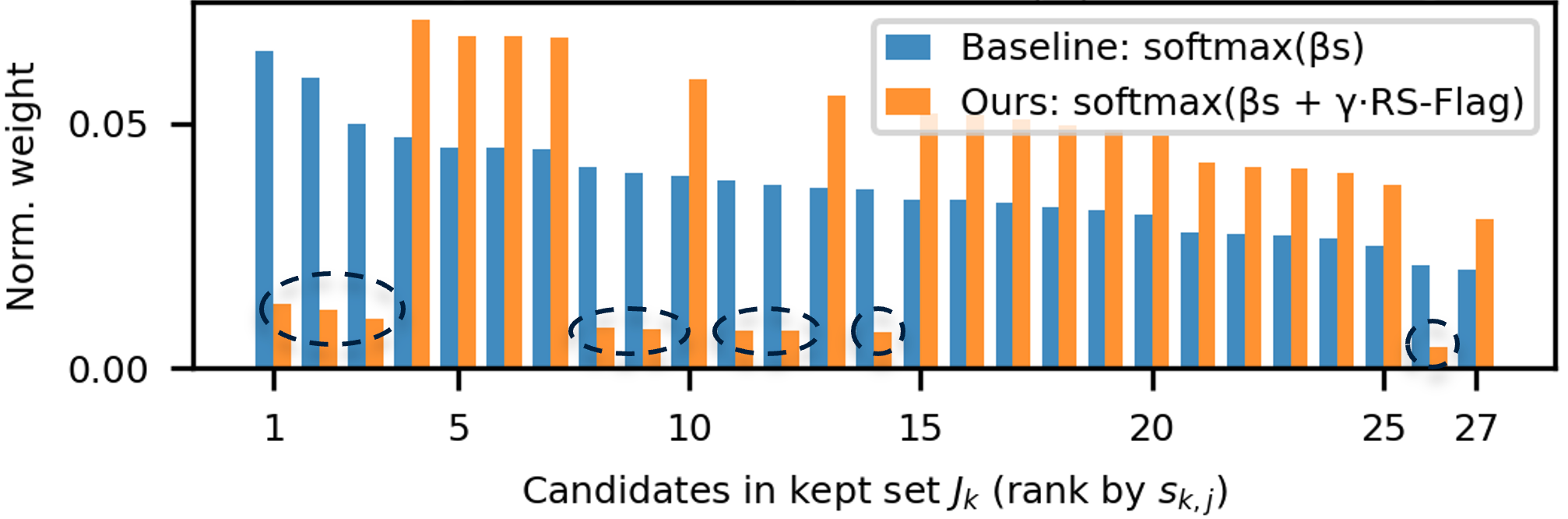}
    \caption{Qualitative analysis of the RS-Flag mechanism using the ``stadium'' class. Candidates are ranked by their initial similarity ($s_{k,j}$). The visualization compares the baseline softmax weights with our calibrated weights. Dashed circles highlight candidates (e.g., indices 1--3, 8--9, 11--12, 14, 26) that lack explicit remote sensing context and are consequently down-weighted to emphasize true RS-aware descriptions.}
    \label{fig:rs_flag_vis}
\end{figure}

To empirically justify the reliability of the RS-Flag mechanism and demonstrate how it mitigates hallucinations, we use the ``stadium'' class as an example to visualize the normalized weights $w$ obtained from Eq.~(3), as shown in Figure~\ref{fig:rs_flag_vis}. The candidates along the x-axis are ranked by their initial visual-textual similarity scores (blue bars). 

As highlighted by the dashed circles, several candidates---most notably the top-ranked ones (indices 1, 2, and 3)---are significantly suppressed after applying the RS-Flag calibration (orange bars). This occurs because these candidates, despite having high initial visual alignment with the image, lack explicit remote sensing contextual features or contain ground-level negative words (i.e., RS-Flag=0). Taking index 3 as an example, a description like ``A massive concrete ring structure surrounding a rectangular green field.'' is not inherently incorrect visually, but it lacks an explicit aerial perspective. Therefore, our mechanism does not directly eliminate it, but instead reduces its relative weight. This calibration allows candidates with stronger RS-awareness (e.g., indices 4, 5, 6, and 7), which receive a positive logit shift (RS-Flag=1), to correctly dominate the final aggregated prototype.

\section{Cross-dataset Evaluation}
\label{sec:cross_dataset}

To further validate the generalization capabilities of AVION and verify that it reduces source-domain overfitting, we conduct a cross-dataset evaluation. We train AVION under a 16-shot setting (using all classes) on the PatternNet dataset and report the zero-shot transfer performance on three target datasets: RESISC-45, UCMerced, and AID (all classes). For fairness, we re-implement all evaluated methods on the identical GeoRSCLIP (ViT-B/32) backbone.

As demonstrated in Table~\ref{tab:cross_dataset}, AVION significantly outperforms APPLeNet on the source domain (+5.46\%) and consistently transfers better to unseen target domains: +2.88\% on RESISC-45, +1.87\% on UCMerced, and +1.59\% on AID. Overall, AVION generalizes better across diverse remote sensing datasets, suggesting that our framework effectively mitigates the common issue of source-domain overfitting.

\begin{table}[ht]
\centering
\caption{Cross-dataset evaluation (\%). Models are trained on PatternNet (16-shot) and evaluated zero-shot on targets. All methods utilize the GeoRSCLIP (ViT-B/32) backbone. ZS: Zero-shot.}
\label{tab:cross_dataset}
\small
\setlength{\tabcolsep}{4pt}
\begin{tabular}{l c ccc}
\toprule
\multirow{2}{*}{Method} & Source & \multicolumn{3}{c}{Target} \\
\cmidrule(lr){2-2} \cmidrule(lr){3-5}
& PatternNet & RESISC-45 & UCMerced & AID \\
\midrule
GeoRSCLIP (ZS)        & 60.17 & 71.89 & 79.72 & 73.72 \\
APPLeNet              & 86.63 & 73.25 & 80.18 & 76.40 \\
\textbf{AVION (Ours)} & \textbf{92.09} & \textbf{76.13} & \textbf{82.05} & \textbf{77.99} \\
\bottomrule
\end{tabular}
\end{table}

\section{Ablation on LLM Domain Prompting}
\label{sec:llm_ablation}

To ensure that the AVION framework is not overly sensitive to the specific large language model employed for domain prompting, we conduct an ablation study evaluating different offline LLM generators. Table~\ref{tab:llm_ablation} reports the average few-shot and base-to-novel generalization results across all six remote sensing datasets using text candidates generated by GPT-5, Llama-3.1-70B-Instruct, and our default Gemini 2.5 Flash.

The performance differences are marginal across all metrics. This stability confirms that AVION's performance gains stem fundamentally from the architectural design of the verification and distillation mechanisms (e.g., RS-Flag and tri-aspect alignment), rather than relying on the idiosyncrasies of a specific language model.

\begin{table}[ht]
\centering
\caption{Ablation on offline LLM choices. We report average few-shot accuracy and base-to-novel generalization (Base, Novel, HM) in \% across six RS datasets. The framework remains stable regardless of the LLM generator used.}
\label{tab:llm_ablation}
\small
\setlength{\tabcolsep}{3.5pt}
\begin{tabular}{l cc ccc}
\toprule
\multirow{2}{*}{Offline LLM} & \multicolumn{2}{c}{Few-Shot} & \multicolumn{3}{c}{Base-to-Novel} \\
\cmidrule(lr){2-3} \cmidrule(lr){4-6}
& 4-shot & 16-shot & Base & Novel & HM \\
\midrule
GPT-5 & \textbf{88.45} & 92.85 & 95.15 & 78.45 & 86.00 \\
Llama-3.1-70B-Inst. & 87.12 & 93.65 & \textbf{95.72} & 79.55 & 86.89 \\
Gemini 2.5 Flash (Ours) & 88.31 & \textbf{93.69} & 95.64 & \textbf{79.94} & \textbf{87.05} \\
\bottomrule
\end{tabular}
\end{table}

\section{Comparisons on Broader Benchmarks}
\label{sec:imagenet}

While AVION is explicitly tailored to address the unique challenges of aerial and satellite imagery, we also explore its scalability and general applicability on the broader, general-domain ImageNet benchmark. 

For this experiment, we evaluate the base-to-novel generalization setting using a standard CLIP (ViT-B/16) student model. To adapt our pipeline to general-domain images, we modify the offline LLM generator query to a general template: ``\textit{In one sentence, describe the distinctive appearance of [CLASS].}'' We generate 30 candidate descriptions per class using Gemini 2.5 Flash. Additionally, because ImageNet consists primarily of ground-level, human-centric photographs, we disable the RS-Flag filtering mechanism. Instead, the framework relies solely on the visual-textual similarity (Eq.~(2)) and the MAD-based robust pruning for selective prototype aggregation.

As shown in Table~\ref{tab:imagenet}, our offline-teacher-to-student distillation paradigm remains highly effective outside the remote sensing domain. When guided by a ViT-H/14 CLIP teacher, AVION outperforms the recent state-of-the-art general prompt tuning method MMRL in terms of the Harmonic Mean (HM). This suggests that the core mechanism of distilling semantically rich, visually verified prototypes is a versatile approach that generalizes well beyond aerial contexts.

\begin{table}[ht]
\centering
\caption{Base-to-novel generalization (\%) on the ImageNet benchmark. The student model is CLIP (ViT-B/16). AVION demonstrates competitive transferability on general-domain images without the RS-Flag constraint.}
\label{tab:imagenet}
\small
\setlength{\tabcolsep}{8pt}
\begin{tabular}{l ccc}
\toprule
Method & Base & Novel & HM \\
\midrule
MMRL & 77.90 & 71.30 & 74.45 \\
\midrule
AVION (Teacher: ViT-L/14) & 77.27 & 74.02 & 75.63 \\
\textbf{AVION (Teacher: ViT-H/14)} & \textbf{79.15} & \textbf{74.79} & \textbf{76.94} \\
\bottomrule
\end{tabular}
\end{table}
\end{document}